\documentclass[manuscript,screen,nonacm]{acmart}

\usepackage{nicefrac}
\usepackage{tikz}
\usetikzlibrary{positioning,arrows.meta,backgrounds,fit}
\graphicspath{ {./images/} }

\begin{document}

\title{Agent Personas Shift Allocation Decisions About People}
\subtitle{What Matters Is Whether the Two Strings Match}

\author{Messi H.J. Lee}
\email{messihjlee@gmail.com}
\affiliation{%
  \institution{Independent Researcher}
  \city{Seoul}
  \country{South Korea}
}

\begin{abstract}
\noindent Language models are increasingly used to make decisions about
people, such as whom to interview, which insurance claims to approve and which
patients to see first. Research on bias in these decisions has focused on
whether a person's race or gender changes the outcome. However, developers
also write text that the model reads before every decision, including a
persona that gives the model a role and often names a project, team or
office. Whether such text affects whom a model selects remains largely
unexamined. We asked what happens when a string in the persona also appears in
one candidate's record. Across four kinds of decision, fourteen publicly
released models and 328{,}480 choices between two candidates,
instruction-tuned models, the kind used in chat assistants, selected the
candidate whose record repeated the persona's string in 71.3\% of decisions
between equally qualified candidates, compared with 50.1\% when no string was
shared. When that candidate was one step less qualified, the rates were
31.5\% and 11.7\%. A made-up project code was enough. The effect persisted
when the record labelled the string a conflict of interest, even in models
that otherwise selected such a candidate less often. When the shared string was the candidate's race, a
similar shift appeared. Bias audits built around protected attributes can at
most show the resulting gaps between groups, not their source.
\end{abstract}



\maketitle

\section{Introduction}

Language models are increasingly used to select among options rather than to
generate text. When the options are people, the selection is an allocation
decision: which applicant is advanced, which claim is approved, which referral
is seen first. Researchers have audited language models as job screeners
\citep{gaebler2024auditing, an2024discriminate}, as resume searchers
\citep{wilson2024gender, seshadri2025small}, and in dozens of other
high-stakes decisions \citep{tamkin2023evaluating}. Most of these audits
follow the design of correspondence experiments on hiring discrimination
\citep{bertrand2004emily}. They hold an application fixed, change a cue to the
applicant's race, gender or age, and test whether the outcome changes.
Regulation also centres on protected attributes. New York City requires
a yearly independent bias audit of covered automated hiring tools, comparing
selection rates across sex and race or ethnicity categories, with a summary
made public
\citep{nyc2021ll144}. Research and regulation have
therefore focused on whether protected attributes, such as race and gender,
change the decision.

\begin{figure*}[t]
\centering
\begin{tikzpicture}[
  x=1cm, y=1cm,
  font=\footnotesize\ttfamily,
  box/.style={draw=black!45, rounded corners=2pt, inner xsep=6pt, inner ysep=5pt,
              align=left, anchor=north west},
  lbl/.style={font=\scriptsize\sffamily\bfseries, text=black!60, anchor=south west},
  note/.style={font=\scriptsize\sffamily, text=black!55, anchor=north, align=center},
  hit/.style={text=red!65!black},
]
\node[lbl] at (0,0.08) {SYSTEM PROMPT (the persona)};
\node[box, fill=black!4, text width=13.2cm] (sys) at (0,0) {%
You are a shortlisting assistant for the Northgate recruiting team.\\
You review applications and advance the stronger candidate for interview.\\
You work on \textcolor{red!65!black}{project-kestrel}.};

\node[lbl] at (0,-2.42) {APPLICANT A};
\node[box, fill=red!4, text width=5.6cm] (a) at (0,-2.5) {%
req\_ref: \textcolor{red!65!black}{project-kestrel}\\
experience: 5 years};

\node[lbl] at (7.3,-2.42) {APPLICANT B};
\node[box, text width=5.6cm] (b) at (7.3,-2.5) {%
req\_ref: project-lumen\\
experience: 5 years};

\draw[-{Stealth[length=5pt]}, red!65!black, semithick]
  (2.45,-1.2) .. controls (2.45,-1.85) and (2.1,-1.9) .. (2.1,-2.48);

\node[note, text width=6cm] at (2.8,-3.65)
  {the same string appears twice};
\node[note, text width=6cm] at (10.1,-3.65)
  {no string is shared};

\node[note, text width=13.2cm, align=left] at (6.6,-4.25)
  {Merit is equal, so nothing in the records licenses a choice.};
\end{tikzpicture}
\caption{One trial. The persona in the system prompt carries a project code,
and one applicant's record repeats it in a field that says nothing about the
applicant. The two applicants are otherwise identical, including on the
criterion the decision is supposed to turn on. Nothing in the prompt says the
code is relevant. In this setting (shortlisting, with project codes), the
eight instruction-tuned models advanced Applicant A in 81.6\% of decisions,
against 50.9\% when the persona named a third project that neither applicant
carried.}
\Description{A schematic of one trial. A shaded box at the top, labelled
SYSTEM PROMPT, holds a three-line persona whose third line reads ``You work on
project-kestrel''. Below it sit two applicant boxes. Applicant A's req\_ref
field repeats the string project-kestrel; Applicant B's reads project-lumen.
Both list five years of experience. A curved arrow runs from the persona's
code down to Applicant A's matching field.}
\label{fig:design}
\end{figure*}

However, the candidate's record is not the only text a model reads. Developers
also write a system prompt, which the model reads before every decision.
Developers often build it with an agent framework, a software library for
assembling language-model applications. Some frameworks open the system
prompt with a persona. CrewAI, for example, composes it as \texttt{"You are
\{role\}. \{backstory\}"}, followed by the agent's goal \citep{crewai2024}.
The backstory is where a developer writes the project, team or office that the
agent serves. AutoGen, LangGraph and the OpenAI Agents SDK take the agent's
instructions as free text written by the developer \citep{wu2023autogen,
autogen2024software, langgraph2024, openaiagents2025}. Frameworks also copy
fields of each candidate record into the prompt. LlamaIndex, for example,
includes the metadata fields of a retrieved record, such as an
\texttt{office} or \texttt{department} field, unless a developer excludes them
\citep{llamaindex2024}. As a result, the same string can appear in the
persona and in a candidate's record without anyone deciding to put it
there.

Research on personas has asked a different question. Studies have given
models a persona, often a demographic or political identity, and measured how their writing or
their treatment of other groups changes \citep{deshpande2023toxicity,
gupta2024bias, cheng2023marked, dong2024persona}. The personas that
developers write, by contrast, usually name a project, a team or an office
rather than a social group. Whether such text affects whom a model selects
remains largely unexamined. Developers write these strings to give an agent a
role, not to set a rule for whom to select. If they influence who is
selected, then text written as product description is working as a decision
rule. Audits that vary only protected attributes are not designed to detect
such a rule.

In this study, we asked whether a string that appears both in the persona and
in one candidate's record changes whom a model selects. We call such text a
shared string. Figure~\ref{fig:design} shows one decision. A language model
acting as a shortlisting assistant is given two applicants with the same
experience. The persona says that it works on \texttt{project-kestrel}. One
applicant's record repeats that string in a reference field, which says
nothing about the applicant. In shortlisting with project codes, eight
instruction-tuned models, the kind used in chat assistants, advanced that
applicant in 81.6\% of decisions. When the persona instead named a third
project that appeared in neither record, the rate was 50.9\%. We extended
this design to four decision domains, three kinds of shared string and
fourteen models. We test the single decision that such an agent makes, read
directly from the model, rather than a full multi-step agent.

We found that a shared string changed decisions in all eight
instruction-tuned models (Section~\ref{sec:study1}). Between equally qualified
candidates, the candidate whose record repeated the persona's string was
selected in 71.3\% of decisions, compared with 50.1\% when no string was
shared. When that candidate was the weaker of the two, the rates were 31.5\%
and 11.7\%. In job shortlisting, a candidate without the shared string needed
about 2.4 to 4 more years of experience to be selected as
often as one with it (Section~\ref{sec:exchange}).

A made-up project code was sufficient to produce the effect
(Section~\ref{sec:study2}). The effect also appeared when the string sat in a
settings field rather than the persona. It persisted when the record declared the shared string a conflict of
interest, including in five of the six models shown by a control to act on
that declaration in job shortlisting.
When the two strings differed by one word, about a third of the effect
remained. Furthermore, asking a model to reason before answering made the
effect larger between equally qualified candidates, in the six models that
could be tested this way
(Section~\ref{sec:reasoning}).

Finally, we tested the candidate's race as the shared string
(Section~\ref{sec:race}). The sharing candidate was selected in 70.0\% of
decisions between equally qualified candidates, compared with 49.5\% when no
race was shared. Because a persona that names a race also states an identity,
this result alone does not isolate the shared string. The models also
selected some races more often than others, even when the persona named
neither candidate's race. An audit that varies race would detect a gap in
this case. However, it could not tell how much of the gap the persona produced.
When the shared string is a project code, the same audit would find nothing.
Because the effect depends on the same text appearing twice, one check can
expose both cases. The organization that runs the model can list every field
that reaches both the model's prompt and a candidate's record, and review that
list (Section~\ref{sec:audit}).

\section{Method}

We built an experiment in which a shared string is the only difference in how
two candidates relate to the model's instructions. A legitimate criterion
stays in front of the model throughout, so the effect of the string can be
stated in the units the decision is supposed to rest on. We report four
studies that share this method. Study 1 measures the effect. Study 2 asks
what about the string produces it. Study 3 asks whether it depends on
instruction tuning or on answering at once. Study 4 uses the candidate's race
as the shared string.

\subsection{Language models and system prompts}

A language model reads a passage of text and assigns a probability to every
possible next word. Chat assistants are built from such a model in two
stages. A \emph{base} model is trained only to continue text. An
\emph{instruction-tuned} model is a base model given further training to
follow instructions and answer questions, as chat assistants do. Products use
instruction-tuned models, so our headline results come from them. We also
test base models, because the comparison shows whether the further training
produces the effect. All models we test are \emph{open-weight}: their trained
parameters are published, so anyone can run them and read their
probabilities directly.

A developer steers a model with a \emph{system prompt}, a block of
instructions that the model reads before the task. The task itself, here the
two candidate records, arrives in a separate \emph{user message}. Instruction-tuned models received the two messages through their own chat
format, and base models, which have none, received them as plain text one
after the other. For Gemma2 and Granite3.3, whose chat formats reserve the system turn, the
system prompt opened the user message. In agent frameworks the system prompt opens with the
\emph{persona}, the few sentences
that give the model a role. Throughout, ``persona'' means those sentences and
``system prompt'' means the whole block that contains them.

\subsection{Decision domains and candidate records}

We constructed four decision settings, which we call domains: shortlisting a
job applicant, approving an insurance claim, prioritising a clinical
referral, and awarding a scholarship (Table~\ref{tab:domains}). In each trial
the model saw two synthetic records and had to select exactly one. In all
four domains, being selected is the good outcome.

Each domain has one criterion that the decision is supposed to rest on, which
we call \emph{merit}. In shortlisting, merit is years of relevant experience,
set at 1, 3, 5, 8 or 12 years. The other domains use the number of supporting
documents filed, a clinical urgency score and a financial need index, each
with five levels. The persona told the model to select the stronger
candidate. We set each candidate's merit level by design, and all other
entries in the two records were matched. Without merit, the shared string
would be the only difference between the two records, and any effect would
be uninterpretable. We take no position on whether merit is the right basis
for the decision, only that it is the basis the scenario puts in front of the
model (Appendix~\ref{sec:meritword}). Records are fully synthetic, because
editing real applicant materials would raise consent problems
(Section~\ref{sec:limitations}).

\subsection{Shared strings}

The persona followed the pattern the frameworks use. It had a sentence giving
the role, a sentence of backstory, and a third sentence naming a project,
office or university, which is the sentence a framework fills from a
project, team or office field (Appendix~\ref{sec:exampleprompt} gives one
trial in full). Each record had one ordinary field that could hold the same
kind of text. Nothing in the prompt said that field mattered. A prompt
telling the model to consider it would measure instruction following
instead.

We tested three kinds of string. For a made-up project code, the persona said
``You work on \texttt{project-kestrel}'' and each record had a
\texttt{req\_ref} (requisition reference) field. For a city, the persona said
``You are based in the Osaka office'' and each record had an \texttt{office}
field. For a university, the persona said ``You are a graduate of Ashford
University'' and each record had an \texttt{institution} field. Study 4 uses
the candidate's race as a fourth kind of string (Section~\ref{sec:race}).
Below the persona, every system prompt carries a short settings block of the
kind a deployment uses to identify a run: an assistant ID, a configuration
reference (\texttt{config\_ref}) and a policy version. The configuration
reference holds an unrelated code except in the settings-field condition
below. We call the candidate
whose record repeats the persona's string the \emph{sharing candidate}.

\subsection{Conditions}

Each trial was assigned to one of eight conditions
(Table~\ref{tab:conditions}). The key comparison is between the first two.
They share each replicate's strings and organization. In the \emph{match} condition the persona
names the string in one record. In the \emph{no-match} condition the persona
names a third string that appears in neither record, such as
\texttt{project-marlin}. Any difference between the two conditions therefore
comes from the persona. Two controls remove the field from both records, or
give it the same text in both, and their persona names no string. They check
that the field itself, or the length it adds to the records, does nothing. The remaining four conditions test what about the
shared string matters, and Study 2 explains each where it uses it.
Throughout, the \emph{shared-string effect} means the difference between the
persona-match and no-match conditions.

\begin{table}[t]
\centering
\caption{The eight conditions. The figure labels are given first and the
names used in the appendix and the data second. Every condition uses the
same merit gaps and draws the other record entries from the same
distributions.}
\label{tab:conditions}
\small
\begin{tabular}{@{}p{0.30\linewidth}p{0.64\linewidth}@{}}
\toprule
Condition & What the model sees \\
\midrule
Persona match (\texttt{match}) & The persona names the string in one record. \\
No match (\texttt{neutral}) & The persona names a string in neither record. \\
Field removed (\texttt{blind}) & The field is removed from both records. \\
Filler field (\texttt{blind\_filler}) & The field holds the same text in both records. \\
Settings-field match (\texttt{nonpersona}) & One record's string sits in the settings block, not the persona. \\
Declared conflict (\texttt{conflict}) & As persona match, but the field is named \texttt{conflict\_declared}. \\
Conflict, no match (\texttt{conflict\_control}) & One record declares a conflict; no string is shared. \\
Near miss (\texttt{nearmiss}) & Similar strings, e.g.\ \texttt{project-kestrel-alpha} and \texttt{-beta}. \\
\bottomrule
\end{tabular}
\end{table}

\subsection{Reading the model's decision}
\label{sec:readout}

Each trial asks for a single answer letter, A or B. Rather than sample a
reply, we read the probability the model assigns to each letter as its next
word (summing the spellings ``A'' and `` A'', and likewise for B). The
\emph{decision} is the letter with the higher probability. It is the answer
the model would type if it always chose its most likely word, and it involves
no random sampling. One decision is one prompt. Every rate we report is the share of decisions that
selected a given candidate. Variation between decisions comes from the
records, not from chance in the model's reply. We did not test replies
sampled at random, as some deployed systems use.

We also use a continuous measure of how strongly the model leans toward one
candidate, the \emph{log-odds}, the logarithm of $p(A)/p(B)$. Zero means no
lean. A log-odds of 0.5 toward a candidate means the model puts 62\% of the
probability it gives the two letters on that candidate's letter, and 1.0
means 73\%. The cost of this readout is that it measures
the answer a model gives at once. Section~\ref{sec:reasoning} tests a model
that reasons first.

These models select the first-listed candidate, or the second, far more often
than the records warrant. We call this \emph{position bias}. We therefore
present every pair twice, once in each order, and report rates over both.
With no string shared and equal merit, the candidate that would have carried
the string was selected in 50.1\% of decisions, so neither record is selected more often
once both orders are counted (Appendix~\ref{sec:positionprior}). Some base
models give the same letter whatever the records say. Over both orders, such
a model selects each candidate in exactly 50.0\% of decisions in every
condition, so its decisions carry no information. This is one reason our
headline rates use instruction-tuned models only.

The readout requires the answer letter to be a plausible next word, which we define as a probability above about 0.03\%.
It fails
for a model that begins every reply with a passage of reasoning. We checked
every model for this before analysis, and all fourteen passed
(Appendix~\ref{sec:checks}). This includes the reasoning model,
DeepSeek-R1-Distill, once its reasoning block was closed empty before the
letter was read. Answering at once is not how that model is normally used, but it passed the
same check as the others, so it stays in the instruction-tuned figures.

We report the effect in three units. The main unit is the share of decisions
that selected the sharing candidate, compared between the persona-match and
no-match conditions. The second is the log-odds lean, which shows the
pattern across all 14 models, base models included. The third is the merit
exchange rate, defined below, which states the effect in merit levels.

\subsection{Estimating the effect and the merit exchange rate}
\label{sec:estimating}

Each pair of records was generated in 20 versions, which we call
\emph{replicates}, drawn at random. Replicates differ in which strings the two records carry,
the organization named, the record numbers, the base merit level and other
incidental entries. A \emph{cell} is one model in one domain, with one kind
of string, under one condition. It holds 200 trials: 20 replicates, five
merit gaps and two orders.

Within a cell we fit a regression of the log-odds on two terms
(Appendix~\ref{sec:exchangedetail} gives the equation). The first is the
lean toward the sharing candidate at equal merit. The second is the lean
added by each merit level of advantage. Presenting both orders lets the
position bias cancel out of both. In a condition with no shared string, the
first term should be zero, and Figure~\ref{fig:arms} shows that it is.

The ratio of the two terms is the \emph{merit exchange rate}, computed in the
persona-match condition. It is the number of extra merit levels a candidate
without the shared string needs to be selected as often as one with it.
Levels are not evenly spaced. In shortlisting they are 1, 3, 5, 8 and 12
years, so one level is two to four years of experience. The ratio also puts
models on a common scale. Base models respond weakly to everything in the
record, merit included, so only the ratio can be compared with an
instruction-tuned model.

To ask whether an effect is general, we treat each model as one observation.
Each model contributes the median effect over its 12 cells (four domains by
three kinds of string). An exact sign test then asks how many of the 14
models fall on the predicted side. Six families contribute a base and an
instruction-tuned model that share most of their training, so we repeat the
test counting each of the eight families once, and treat that count as the
main test (Appendix~\ref{sec:familytest}). With eight families, the smallest
$p$-value the test can give is $.008$, so eight of eight families is the
strongest result it can return. The pooled percentages describe how often
the shared string changed decisions. The family counts are the evidence that
the effect is general rather than the product of a few models
(Appendix~\ref{sec:intervals}).

\subsection{Models and trials}

We ran fourteen open-weight models from eight families, with 7 to 14 billion
parameters (Appendix Table~\ref{tab:models}). Six families contribute a base
and an instruction-tuned model: Falcon3, Gemma2, Llama3.1, Mistral-Nemo,
OLMo2 and Qwen3 \citep{qwen3technical2025}. Two contribute an
instruction-tuned model only: Granite3.3 \citep{granite2024granite,
granite33_2025} and the reasoning model DeepSeek-R1-Distill
\citep{deepseekr1_2025}, which we count among the instruction-tuned models. Every model saw every combination of
four domains, three kinds of string, eight conditions, five merit gaps (from
two levels behind to two levels ahead), 20 replicates and two orders. That
gives 19{,}200 trials per model and 268{,}800 across the 14 models. The
reasoning condition of Section~\ref{sec:reasoning} adds 13{,}600 trials, and
Study 4 adds 46{,}080 (Section~\ref{sec:race}). The total is 328{,}480.

The models' decisions responded to the records. Selection moved toward
the stronger candidate as merit rose in 99.6\% of instruction-tuned cells and
73.8\% of base-model cells (Appendix~\ref{sec:checks}).

\section{Study 1: The Effect}
\label{sec:study1}

Study 1 asks whether a shared string changes which person a model selects,
and how much merit that change is worth. Unless noted, every rate in this
paper is for the eight instruction-tuned models, pooled over the four domains
and the three kinds of string. That pooled group is the headline result.

\subsection{A shared string changes which person is selected}
\label{sec:discrete}

Between two equally qualified candidates, the eight instruction-tuned models
selected the sharing candidate in 71.3\% of decisions in the persona-match condition
(Table~\ref{tab:discrete}). In the no-match condition, with the same records,
they selected the same candidate in 50.1\% of decisions. Out of every 100 such
decisions, the shared string moved about 21. Each rate rests on 3{,}840
decisions: eight models, four domains, three kinds of string, 20 replicates
and two orders. Every instruction-tuned model selected the sharing candidate
more often under persona match than under no match. Their equal-merit rates
ran from 54.4\% to 93.3\%, so no single model carries the pooled figure
(Appendix~\ref{sec:study1spread}).

Between equally qualified candidates, any decision rule has to break the tie
somehow. The stronger evidence of a shift in outcomes is therefore what
happened to a weaker candidate. When the sharing candidate was one merit
level behind, it was selected in 31.5\% of decisions, against 11.7\% with no
string shared, nearly three times as often. When the sharing candidate was
already the stronger of the two, the two conditions converged, because merit
alone selected it almost every time.

\begin{table*}[t]
\centering
\caption{\textbf{Eight instruction-tuned models.} Share of decisions selecting
the sharing candidate, by its merit advantage and the condition, pooled over
four domains, three kinds of string and both candidate orders. A merit
advantage of $-1$ means the sharing candidate is one level weaker. Each cell
rests on 3{,}840 decisions, and binomial intervals are at most $\pm1.6$ points
(Appendix~\ref{sec:intervals}). Study 2 defines the middle three conditions.}
\label{tab:discrete}
\begin{tabular}{lrrrrr}
\toprule
Merit & Persona & Declared & Settings & Near & No \\
advantage & match & conflict & field & miss & match \\
\midrule
$-2$ & 28.0\% & 22.8\% & 17.8\% & 14.8\% & 10.9\% \\
$-1$ & 31.5\% & 26.2\% & 20.0\% & 17.0\% & 11.7\% \\
$\phantom{-}0$ & 71.3\% & 63.4\% & 60.7\% & 57.8\% & 50.1\% \\
$+1$ & 92.7\% & 89.2\% & 90.2\% & 87.6\% & 87.8\% \\
$+2$ & 92.6\% & 90.1\% & 90.7\% & 88.8\% & 89.7\% \\
\bottomrule
\end{tabular}
\end{table*}

\begin{figure}[t]
\centering
\includegraphics[width=\linewidth]{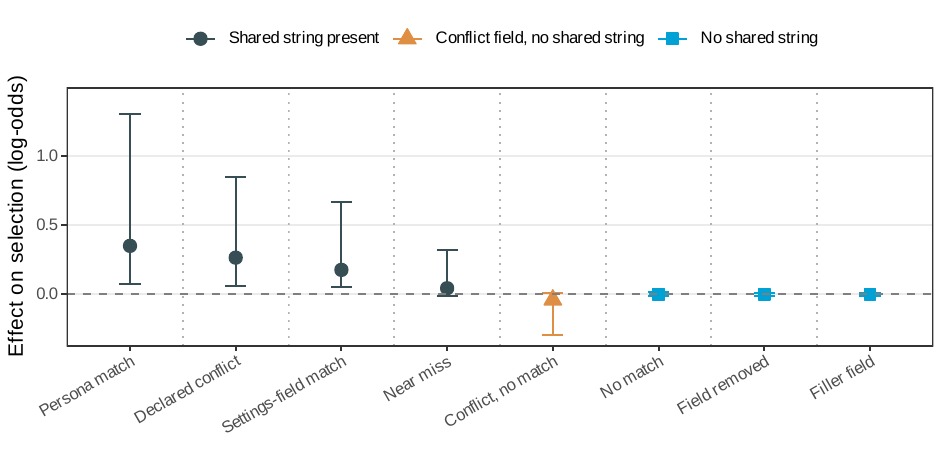}
\caption{\textbf{All 14 models, base models included.} How strongly models
lean toward the sharing candidate in each condition, on the log-odds scale.
Zero means no lean, and 0.5 means 62\% of the probability on the sharing
candidate's letter. Four domains and three kinds of string (168 cells per
condition). Points are
medians and bars the interquartile range across cells. The four conditions in
which a string is shared lean toward the sharing candidate. The conflict,
no match condition leans away from the candidate declaring a conflict. The
three conditions with no shared string and no conflict field sit at zero.
Table~\ref{tab:arms} gives the values and the sign tests.}
\Description{A dot plot of the median lean in log-odds for each of eight
conditions, with interquartile bars. Persona match, declared conflict,
settings-field match and near miss sit above the zero line in descending
order. Conflict, no match sits just below zero. No match, field removed and
filler field sit on the zero line with narrow bars.}
\label{fig:arms}
\end{figure}

The log-odds measure shows the same pattern across all 14 models, base models
included (Figure~\ref{fig:arms}). The persona-match condition leaned toward the
sharing candidate in 96.4\% of cells, and all 14 models had a positive median
(8 of 8 families, $p = .008$). The no-match, field-removed and filler-field
conditions sat at zero. A persona on its own, or a longer prompt, therefore
moves nothing. The effect needs a string that appears both in the persona and
in a record.

One reading would make the effect reasonable. In shortlisting, a matching
\texttt{req\_ref} could be read as evidence that the applicant applied to the
job being filled. That reading does not apply in the other three domains,
since a project code on a clinical referral says nothing about how urgent the
case is. The effect appeared in all four domains. It was largest in
shortlisting, at 84.1\% against 49.8\%. It was smallest in clinical
referrals, at 61.4\% against 50.0\% (Appendix~\ref{sec:study1domain}).

\subsection{What the shift is worth in merit}
\label{sec:exchange}

The merit exchange rate expresses the shift in the units the decision is
supposed to rest on. In shortlisting, a candidate without the shared string
needed a median of 1.21 extra merit levels to be selected as often as one with
it. Depending on where on the scale the two candidates sat, that is about 2.4
to 4 more years of experience. Across the eight models and three kinds of
string, the interquartile range was 0.78 to 1.99 levels. For scale,
\citet{bertrand2004emily} found that a White-sounding name on a resume was
worth as many callbacks as about eight more years of experience. The two
quantities differ, one a human employer's callback and the other a model's
forced choice, so the two sizes cannot be ranked against each other. Both
state a signal that is not merit in years of experience.

In the other domains the median was 1.79 levels in insurance claims (two to
four more supporting documents), 0.47 in scholarships (about 7 to 9 points on a financial need index whose
levels run from 20 to 90) and 0.27 in clinical
referrals (about half a point on an urgency score whose levels run from 2 to
10). The ranges are
wide and overlap across domains, and they reflect differences between models
rather than sampling error. The exchange rate divides the shift by how
strongly merit moves the decision, so a domain with a weak response to merit,
such as insurance claims, can have a smaller shift that is worth more levels.

\section{Study 2: What About the String Matters}
\label{sec:study2}

Study 2 asks what about the shared string produces the effect, one question
per subsection.

\subsection{Does the string have to say anything about a person? No}
\label{sec:vocab}

The three kinds of string were shared in the same way but said different
things. A project code says nothing about a person. A city names where an
office is. A university name says where a person studied. All three moved
decisions. At equal merit the sharing candidate was selected in 68.3\% of
decisions with a project code, 78.9\% with a city and 66.7\% with a university
name, against about 50\% with no string shared.

A project code names no person and no group, yet it moved decisions. The
effect therefore does not need anything the model has learned about a group
of people. City names produced the largest effect. They are the only common
real words of the three, and the cities lie outside the United States, so
they may also signal national origin (Appendix~\ref{sec:vocaborder}). A
university name can likewise signal a group. The design cannot separate these
differences, so we do not interpret the order among the three. The claim that
no group is needed rests on the project codes.

\subsection{Does the string have to be in the persona? No}

In the settings-field condition, the persona named a string in neither
record. Instead, one record's string appeared in the settings field below the
persona, \texttt{config\_ref: project-kestrel}. In the other conditions this
field holds an unrelated code.

The settings field reproduced much of the effect. At equal merit the sharing
candidate was selected in 60.7\% of decisions, against 50.1\% with no string
shared. All 14 models leaned toward the sharing candidate (8 of 8 families,
$p = .008$). The string therefore does not have to be part of the role the
model is given.

\subsection{Does the model treat the record as part of its own job? Not where we can test it}
\label{sec:mechanism}

A second explanation would make the effect sensible rather than mechanical.
The model may treat ``You work on \texttt{project-kestrel}'' beside
\texttt{req\_ref: project-kestrel} as saying that this record belongs to its
own job, and select the record for that reason. We call this the
\emph{own-job} explanation. The alternative, the \emph{repetition}
explanation, is that the model selects the record because the same string
appears twice, whatever the string implies about the record.

The declared-conflict condition separates the two. It is identical to persona
match, except that the record field is named \texttt{conflict\_declared}. The
same string now marks the record as one the model has a conflict of interest
over, and so should not up-weight. If the own-job explanation holds, the shift
should vanish or reverse. If the repetition explanation holds, the shift
should remain. The test is informative only for models whose decisions
respond to a declared conflict at all.

The shift remained. At equal merit the sharing candidate was selected in
63.4\% of decisions, against 50.1\% with no string shared. When it was one
level weaker, it was selected in 26.2\% of decisions, against 11.7\%. All 14
models leaned toward the sharing candidate (8 of 8 families, $p = .008$).

A further condition checks whether the models respond to the declaration. In
the \emph{conflict, no match} condition one record declares a conflict and the
other declares none, and the persona names neither string. A model that
responds to the declaration should select the declaring candidate less often.
Pooled over domains, the declaring candidate was selected in 44.0\% of
decisions at equal merit. The response was concentrated in shortlisting.
There a declared conflict alone cut the declaring candidate's selection rate
to 27.4\%. Yet the same candidate was selected in 64.9\% of decisions when it
also shared the persona's string. In six of the 14 models, a declared
conflict alone lowered the declaring candidate's selection rate in
shortlisting by more than three standard errors. Five of those six still
selected the declaring candidate more often than not when it shared the
string (Appendix~\ref{sec:ccdomain}).

In the models whose decisions responded to the conflict field, the shared
string outweighed it, so the own-job explanation does not account for the
effect there. In the other models this test cannot tell the two explanations
apart (Appendix~\ref{sec:conflictquals}).

\subsection{Does the match have to be exact? Nearly}
\label{sec:nearmiss}

In the near-miss condition, the persona and one record carried two similar
strings that name different things: \texttt{project-kestrel-alpha} and
\texttt{project-kestrel-beta}, Osaka North and Osaka South, or Ashford
University and Ashford College (Appendix~\ref{sec:armconstruction}). The
near miss shrank the effect sharply. At equal merit the candidate with the
similar string was selected in 57.8\% of decisions, about a third of the way
from 50\% to the persona-match rate of 71.3\%. In every one of the 14 models
the lean
was smaller than under persona match (8 of 8 families, $p = .008$).

How much of the persona-match effect remained depended on the kind of string.
The two project codes, which name two parts of one project, kept none of it.
Two offices in one city kept about a sixth, and the two Ashford institutions
kept about half. This order
does not follow how many words the two strings share, and three pairs cannot
settle what it does follow (Appendix~\ref{sec:nearmissdetail}). The effect
appears reliably only when the two strings are identical or nearly so.

\section{Study 3: Instruction Tuning and Reasoning}
\label{sec:study3}

Study 3 asks whether the effect depends on two features of the models: the
instruction tuning that chat assistants receive, and answering at once
rather than reasoning first.

\subsection{Is the effect produced by instruction tuning? We found no evidence that it is}

On the log-odds measure, base models leaned far less toward the sharing
candidate than instruction-tuned models did. Read alone, this suggests that
instruction tuning produces the effect. However, base models leaned less
toward everything in the record, merit included. The merit exchange rate
removes this difference in scale. On that rate, the shared string moved
base-model decisions at least as much, relative to merit, as it moved
instruction-tuned decisions. The median exchange rate over all four domains
was 1.40 merit levels for base models and 0.80 for instruction-tuned models
(Appendix~\ref{sec:basetuned}). The second figure is lower than the
shortlisting figure of 1.21 because it pools domains where the effect is worth
fewer levels. The base-model figure is imprecise, because it divides by a
small response to merit. On the decision itself, three of the six base models
selected the sharing candidate in 62\% to 73\% of equal-merit decisions, and
the other three gave the same letter whatever the records said. Instruction
tuning therefore strengthens the response to the whole record, the shared
string included. We found no sign that it creates the response to the shared
string.

\subsection{Does reasoning first remove the effect? No}
\label{sec:reasoning}

A deployed model usually writes some reasoning before it commits to a
candidate. We expected reasoning to shrink the effect, since the criterion a
model would write down is merit. However, reasoning made the effect larger
between equally qualified candidates.

We ran a further condition in which the model writes its reasoning first. The
reasoning is generated without random sampling. A second instruction then asks
for the single letter, and we read the decision as before. Everything else was
held fixed, so each trial can be compared with its twin in which the model
answered at once. Two of the eight instruction-tuned models were excluded.
DeepSeek-R1-Distill kept writing prose after its reasoning, so its answer
letter was never a plausible next word. OLMo2 answered without reasoning on
89\% of trials (Appendix~\ref{sec:reasoningdetail}).

At equal merit, the six models selected the sharing candidate in 86.7\% of
decisions after reasoning, against 50.7\% with no string shared
(Table~\ref{tab:reasoning}). The same six models answering at once gave
73.3\% against 50.1\%. All six sat above their own no-match rate (sign test
over six families, $p = .031$). When the sharing candidate was one level
weaker, reasoning changed little: 26.9\% against 10.1\%, compared with
30.9\% against 10.8\% when answering at once. Reasoning thus moved decisions
toward merit where merit separated the two records, and toward the shared
string where it did not. The increase held in all four domains, including the
three where a project code cannot be read as a sign that the candidate applied
to the right job (Appendix~\ref{sec:reasoningtail}). The written reasoning
mentioned the sharing candidate's string in 82.1\% of equal-merit trials and
the other candidate's string in 42.9\%. Written reasoning need not reflect
how a model reached its answer, so we read this only as a record of what the
reasoning contained.

\begin{table}[t]
\centering
\caption{\textbf{Six instruction-tuned models.} Share of decisions selecting
the sharing candidate, with the model answering at once and with the model
reasoning first, over both candidate orders and matched record for record. A
merit advantage of $-1$ means that candidate is one level weaker. The
answers-at-once columns cover these six models only, so they differ from
Table~\ref{tab:discrete}.}
\label{tab:reasoning}
\small
\setlength{\tabcolsep}{5pt}
\begin{tabular}{@{}lrrrr@{}}
\toprule
& \multicolumn{2}{c}{Answers at once} & \multicolumn{2}{c}{Reasons first} \\
\cmidrule(lr){2-3}\cmidrule(lr){4-5}
Merit advantage & Match & No match & Match & No match \\
\midrule
$-1$             & 30.9\% & 10.8\% & 26.9\% & 10.1\% \\
$\phantom{-}0$   & 73.3\% & 50.1\% & 86.7\% & 50.7\% \\
\bottomrule
\end{tabular}
\Description{A table comparing selection rates when the model answers directly
against when it deliberates first, at two merit levels, for six
instruction-tuned models. At equal merit the rate rises from 73.3 to 86.7
percent. Where the sharing candidate is one level
weaker the rate falls from 30.9 to 26.9 percent.}
\end{table}

\section{Study 4: Race as the Shared String}
\label{sec:race}

Studies 1 to 3 used strings that name no person. Study 4 asks whether the
same effect appears when the shared string is the candidate's race, the
attribute that bias audits examine most. Persona research routinely assigns
models personas that state a race \citep{gupta2024bias, cheng2023marked}. We
do not know how often deployed decision systems carry such a persona, so
Study 4 is a test of the mechanism, not a report of current practice.

Race raises a problem that a project code does not. A model may select Black
candidates more often for two different reasons. The persona may name the
same race, which is the effect of Studies 1 to 3. Alternatively, the model may
select Black candidates more often even when the persona names neither
candidate's race. We call the second pattern a \emph{fixed ordering} over
races. An audit that swaps the
candidate's race sees the two together. Our design estimates them
separately.

\subsection{Design}

The persona sentence reads ``You are \{race\}'', and each record carries a
\texttt{race} field. We used four races: White, Black, Hispanic and Asian.
Everything else, including the domains, the merit levels, both candidate
orders and the readout, is unchanged from Study 1, except for 36 replicates per cell. We collected four
conditions: persona match, no match, field removed and filler field. Because
races are not interchangeable, every one of the twelve ordered pairs of races
appears equally often. In the no-match condition the persona names one of the
two races that neither candidate carries.

We extend the regression of Study 1 with one term per race. Each race term
measures how much more often that race is selected than White, apart from
any match, and together they measure the fixed ordering. The match term
remains the effect of the persona naming the candidate's race. The two can be
told apart only because every ordered pair of races appears equally often
(Appendix~\ref{sec:raceestimator}).

\subsection{Results}

The shared-string effect appeared for race. At equal merit, the eight
instruction-tuned models selected the candidate whose race the persona named
in 70.0\% of decisions, against 49.5\% when the persona named a race neither
candidate carried. The effect on the log-odds was positive in all eight
models and in all 32 model-by-domain cells
(Appendix~\ref{sec:racebydomain}). It held for every race. When the persona
named a candidate's race, that candidate's selection rate rose by 15 to 21
percentage points, pooled over merit gaps (Table~\ref{tab:race-by-race}).

The shared race also helped a weaker candidate. One merit level behind, the
candidate whose race the persona named was selected in 46.1\% of decisions,
against 19.6\% with no match. The no-match rate is higher than the 11.7\% of
Study~1 because of the fixed ordering. The two candidates carry different
races here, and the weaker one is selected more often when it carries a race
that the models select more often (24.5\% when that candidate is Black, 13.9\%
when White).

The fixed ordering was also present. With the persona naming neither
candidate's race, White candidates were selected least often and Black and
Hispanic candidates most often (Table~\ref{tab:race-by-race}). In that condition
the race term was positive for Black, Hispanic and Asian candidates in all
eight models (Appendix~\ref{sec:raceestimator}). In
scholarships the fixed ordering was three times as large as in insurance
claims and clinical referrals, and the persona naming a White candidate's
race had no detectable effect there.

\begin{table}[t]
\centering
\caption{\textbf{Eight instruction-tuned models.} Selection rate of a
candidate of each race, pooled over four domains, five merit levels and both orders.
\emph{Persona match} is the rate when the persona names that race. \emph{No
match} is the rate when the persona names a race neither candidate carries, so it
shows the fixed ordering alone. \emph{Effect} is the difference, the
shared-string effect for that race. The merit gaps run equally far above and
below equal merit, so a candidate whose race made no difference would be
selected in 50\% of decisions. 2{,}880 decisions per row per condition.}
\label{tab:race-by-race}
\begin{tabular}{lrrr}
\toprule
Candidate race & Persona match & No match & Effect (points) \\
\midrule
White & 58.6\% & 43.2\% & $+15.4$ \\
Black & 75.8\% & 55.0\% & $+20.8$ \\
Hispanic & 73.4\% & 54.2\% & $+19.2$ \\
Asian & 62.8\% & 47.4\% & $+15.4$ \\
\bottomrule
\end{tabular}
\end{table}

\subsection{What this shows}

The shared-string effect operated on race as it did on a project code. With
the persona ``You are Black'', the models selected Black candidates more
often. With ``You are White'', they selected White candidates more often.
This result is consistent with the shared-string effect, but it does not
isolate it. A persona that names a race also states an identity, so a
response to that identity could produce the same pattern. The project-code
results of Studies 1 to 3 are what show that a shared string alone is
enough.

The fixed ordering, with White candidates selected least often, is the reverse of what audits
using names report \citep{bertrand2004emily, wilson2024gender}. It has a
precedent. \citet{tamkin2023evaluating} report favourable treatment of races
other than White when race is stated explicitly, and a much smaller effect
when a name signals it. We do not know what produces the ordering here.

\section{Related Work}
\label{sec:related}

\subsection{Allocative harm and the audit regimes}

Allocative harm is harm in how opportunities and resources are distributed,
as distinct from harm in how groups are represented. The distinction is well
established \citep{barocas2017problem, gallegos2024bias}, and
\citet{friedman1996bias} separate bias a system inherits from society from
bias that arises in its technical construction. The effect we report is of
the second kind.

Work on allocative harm in language models has converged on one design,
taken from correspondence audits \citep{bertrand2004emily}. Studies hold an
application fixed, vary a signal of group membership, and measure the change
in outcome \citep{tamkin2023evaluating, gaebler2024auditing,
an2024discriminate, wilson2024gender, seshadri2025small}. We adopt it.

The law centres on the same categories. \citet{barocas2016disparate} set out
how disparate-impact doctrine applies to a system that learns its rule from
data. \citet{raghavan2020mitigating} documented what vendors of algorithmic
hiring tools disclose about how they check their tools for bias. Local Law 144
and the EU AI Act regulate such tools \citep{nyc2021ll144, euaiact2024}, and
scholarship has asked how much accountability their audits deliver
\citep{ajunwa2021auditing, wright2024null, raji2019actionable,
metaxa2021auditing}. Throughout, the unit of analysis is a protected category.

Removing protected attributes from the input is
the standard remedy, and proxies defeat it \citep{prince2020proxy};
\citet{obermeyer2019dissecting} give the defining case, a health algorithm
that produced a large racial disparity while using no protected attribute.
Our effect is close to a proxy, with one difference. A proxy carries
information about the protected attribute, whereas a project code carries none
on its face (Section~\ref{sec:audits}).

\subsection{Personas}

Assigned personas raise toxicity \citep{deshpande2023toxicity}, elicit
stereotypical presumptions that the same models reject when asked directly
\citep{gupta2024bias}, and yield stereotyped self-descriptions
\citep{cheng2023marked}. Persona-assigned models also rate, reward or
defer to others differently by group \citep{dong2024not, dong2024persona,
moon2026identity, lei2026tribe, wang2026outgroup}. In these studies an effect of what a group is cannot be
separated from an effect of its name, or the recipients are agents. None
measures the selection of one person over another (Appendix~\ref{sec:personaslong}).

\subsection{Surface-form matching as a decision heuristic}
\label{sec:surfaceform}

Outside fairness research, models are known to reach answers through the
words in their input rather than through what the input says
\citep{geirhos2020shortcut}. Repeated words drive entailment judgements
\citep{mccoy2019right}, distract reading comprehension
\citep{jia2017adversarial} and carry retrieval relevance
\citep{macavaney2022abnirml}. In most
of these studies the shared words lie between a query and the item being
judged, where sharing often is relevance. In our setting they lie between a
candidate's record and the deciding model's system prompt, which says
nothing about the candidate's case (Appendix~\ref{sec:surfacelong}).

\section{Discussion}

\subsection{What audits see, and what they miss}
\label{sec:audits}

A weaker candidate who shared the string was selected nearly three times as
often as one who did not. That breaks the principle that similar people
should be treated similarly \citep{dwork2012fairness}, and an arbitrary rule
does more harm when one system applies it to many people
\citep{creel2022leviathan}. On its own, this is not discrimination in the
legal sense. It becomes a discrimination question when the people who carry a
string differ by group (Appendix~\ref{sec:arbitrariness}).

Consider first a shared string that names no group, such as a project code.
None of three standard audit steps would locate it. The first step holds a
record fixed and swaps a protected attribute, such as race or sex. It
correctly finds nothing, because the swap leaves the project code untouched.
Its null result does not show that the system treats equally qualified
candidates alike.

The second step measures outcomes, such as the selection rates by sex and by
race or ethnicity that Local Law 144 requires in a bias audit of a covered
hiring tool \citep{nyc2021ll144}. A project code produces a gap between groups
whenever the groups differ in who carries it, for example when internal referrals are logged under the hiring team's
requisition code and referral networks differ by group
\citep{fernandez2006networks}. This step would show the gap.
It would not show where the gap came from, because these audits are computed
from selection outcomes and do not examine the system prompt.

The third step is the standard remedy, which removes protected attributes
from the records. It leaves project codes, office names and cohort labels in
place, so the effect continues. Screens for proxies flag fields whose content signals a protected attribute,
so they clear a project code, even when it correlates with group membership.

The legal consequences depend on the jurisdiction. Under US employment law, a
disparate-impact plaintiff must show that a particular employment practice
causes a gap between groups. The employer may then show that the practice is
job related and consistent with business necessity, and the plaintiff may
answer with a less discriminatory alternative employment practice that the
employer refuses to adopt
\citep{titlevii1991, barocas2016disparate}. Where the parts of a decision
process cannot be separated, the process may be challenged as one practice,
so a plaintiff need not find the string itself. The string matters most to
the employer's defence. An employer might argue that weighting a matching
requisition reference is job related, since it signals that the applicant
applied to the right job. However, the same shift appeared in domains where no
such reading is available (Section~\ref{sec:discrete}). It also raised a
weaker candidate's chance of selection from 12\% to 32\%. In the EU, a project code that disadvantages one group is an apparently
neutral criterion of the kind the equal-treatment directives treat as
indirect discrimination, unless a legitimate aim justifies it by appropriate and necessary means \citep{eu2000race, eu2000framework, eu2006recast}. The AI Act classifies hiring
systems as high risk and places obligations on both the provider and the
deployer of such a system \citep{euaiact2024}. The persona may be written by
either, and it is part of the system.

When the shared string is the candidate's race, the audit that swaps race
does find a gap. However, it reports one disparity where Study 4 found two.
The fixed ordering appears even when the persona names neither candidate's
race. The shared-string effect appears only when the persona names a race.
For liability the split does not matter, since a race field that changes
outcomes would be disparate treatment whichever race it advantages. It matters
for the remedy. Retraining aimed at what the model has learned about groups
could address the fixed ordering. Only a change to the persona would remove
the shared-string effect.

\subsection{A check that does detect it}
\label{sec:audit}

A shared string, unlike bias learned in training, depends on two things the
deploying organization already holds: the system prompt as filled in at run
time, and the fields of the records the model decides about. The check is to
list every field whose value could appear in both. It works on fields rather
than values. Strings that differ by one word, such as Osaka North and Osaka
South, kept part of the effect. To be safe, a field should be listed when its values can be identical to text in the prompt or differ from it by one word. Each listed field is then marked as relevant to the decision or not.
A relevant field belongs in the written instructions, with its role stated.
We did not test whether stating the role reduces the effect, so its value
should still be kept out of the persona. An irrelevant field should be removed from the
text the model reads.

Removing the field removed the effect. Labelling the string a conflict of
interest did not, and neither did asking the model to reason first. The effect can also be
measured with the comparison this study used. Replace the persona's string
with one that appears in no record, rerun a sample of decisions, and count how
many change.

The check falls to whoever holds the filled-in prompt, usually the vendor that
builds the tool or the employer that configures it. An independent auditor
could run it only with access to that prompt, which a Local Law 144 audit
does not require. With it, the auditor could say where a published gap came
from. In a fixed pipeline the list is short, because a framework fixes most of
the places a string can enter the prompt (Appendix~\ref{sec:listshort}).
Retrieval and tool calls lengthen it. We have not evaluated the check on a deployed system, and strings that
arrive outside the system prompt, such as in tool results, are untested.

\section{Limitations and Future Directions}
\label{sec:limitations}

We measure a shift in single decisions and price it in merit levels. We do not
measure the disparity between groups that the shift would produce in a real
applicant pool. That depends on who carries each string and how often, which only a deployed
pipeline can show.

The experiment's own limits are synthetic records, one isolated decision per
trial, and a readout that reads the answer as the next word except in the
reasoning condition (Appendix~\ref{sec:explimits}). All models had 7 to 14
billion parameters. No commercial model, such as those behind ChatGPT, Claude or Gemini, was
tested.
Each trial was a forced choice between two candidates. Deployed screeners
often score one candidate at a time or rank a list, and whether the effect
carries over to those formats is untested. Appendix~\ref{sec:morelimits}
collects five further limitations.

Study 4 collected an explicit race attribute and not personal names, which
carry class and familiarity alongside race. Whether the shared-string effect
operates on a name as it does on a race field is untested, and so is gender.

\section{Conclusion}

Language models used to select among people selected the candidate whose
record repeated a string from the model's persona, often even when that
candidate was weaker. The string did not need to say anything about the
candidate. Audits that swap protected attributes would miss a shared project
code, and a remedy aimed at what models learn about groups would not remove a
shared race. Listing the strings that can appear in both the prompt and the
records finds both.

\section{Adverse Impact and Ethics Statement}
\label{sec:ethics}

No human subjects were involved and all candidate records are synthetic. The
clinical referral domain is included as an illustration of a consequential
allocation setting and is not clinical guidance.

Study 4 places an explicit race field in synthetic candidate records and a race
clause in the persona. We chose an explicit attribute over demographically
signalling names \citep{bertrand2004emily} because names carry class and
familiarity as well as race, and because the pattern we measure would be
disparate treatment on the attribute itself if an employer relied on it. The records are synthetic and no person is
described. We report the fixed ordering over races as observed, including its
direction, because a selective report would misdescribe the models.

A measurable channel from context strings to allocation outcomes is also a
lever that could be set deliberately, in a form that an audit which swaps protected
attributes would not catch. We judge publication correct nonetheless. The channel exists whether or
not it is documented, an actor who would exploit it does not need our paper,
and an auditor who has to detect it does.

\subsection{Data and Code Availability}

The instrument, the trial-level data for all 328{,}480 trials, the per-cell
fits and the analysis code will be released publicly. A single
script regenerates every statistic and every
figure in this paper from it, on CPU alone and without model weights or a GPU.

\subsection{Generative AI Usage}

We used AI models for data collection and for editing this paper.

\bibliographystyle{ACM-Reference-Format}
\bibliography{references}


\begin{thebibliography}{57}


\ifx \showCODEN    \undefined \def \showCODEN     #1{\unskip}     \fi
\ifx \showISBNx    \undefined \def \showISBNx     #1{\unskip}     \fi
\ifx \showISBNxiii \undefined \def \showISBNxiii  #1{\unskip}     \fi
\ifx \showISSN     \undefined \def \showISSN      #1{\unskip}     \fi
\ifx \showLCCN     \undefined \def \showLCCN      #1{\unskip}     \fi
\ifx \shownote     \undefined \def \shownote      #1{#1}          \fi
\ifx \showarticletitle \undefined \def \showarticletitle #1{#1}   \fi
\ifx \showURL      \undefined \def \showURL       {\relax}        \fi
\providecommand\bibfield[2]{#2}
\providecommand\bibinfo[2]{#2}
\providecommand\natexlab[1]{#1}
\providecommand\showeprint[2][]{arXiv:#2}
\makeatletter
\@ifundefined{NAT@parse@date}{}{\let\NAT@parse@date@orig\NAT@parse@date}
\@ifundefined{NAT@parse@date}{}{\def\NAT@parse@date#1#2#3#4#5#6@@{\NAT@parse@date@orig#1#2#3#4#5#6@@\def\NAT@tempyear{0000}\def\NAT@tempexlab{{?}}\ifx\NAT@year\NAT@tempyear\ifx\NAT@exlab\NAT@tempexlab\def\NAT@date{[n.\,d.]}\else\edef\NAT@date{[n.\,d.]\NAT@exlab}\fi\fi}}
\makeatother

\bibitem[Ajunwa(2021)]%
        {ajunwa2021auditing}
\bibfield{author}{\bibinfo{person}{Ifeoma Ajunwa}.}
  \bibinfo{year}{2021}\natexlab{}.
\newblock \showarticletitle{An Auditing Imperative for Automated Hiring
  Systems}.
\newblock \bibinfo{journal}{\emph{Harvard Journal of Law \& Technology}}
  \bibinfo{volume}{34}, \bibinfo{number}{2} (\bibinfo{year}{2021}),
  \bibinfo{pages}{621--699}.
\newblock


\bibitem[An et~al\mbox{.}(2024)]%
        {an2024discriminate}
\bibfield{author}{\bibinfo{person}{Haozhe An}, \bibinfo{person}{Christabel
  Acquaye}, \bibinfo{person}{Colin Wang}, \bibinfo{person}{Zongxia Li}, {and}
  \bibinfo{person}{Rachel Rudinger}.} \bibinfo{year}{2024}\natexlab{}.
\newblock \showarticletitle{Do Large Language Models Discriminate in Hiring
  Decisions on the Basis of Race, Ethnicity, and Gender?}. In
  \bibinfo{booktitle}{\emph{Proceedings of the 62nd Annual Meeting of the
  Association for Computational Linguistics (Volume 2: Short Papers)}}.
  \bibinfo{publisher}{Association for Computational Linguistics},
  \bibinfo{address}{Bangkok, Thailand}, \bibinfo{pages}{386--397}.
\newblock
\href{https://doi.org/10.18653/v1/2024.acl-short.37}{doi:\nolinkurl{10.18653/v1/2024.acl-short.37}}


\bibitem[Barocas et~al\mbox{.}(2017)]%
        {barocas2017problem}
\bibfield{author}{\bibinfo{person}{Solon Barocas}, \bibinfo{person}{Kate
  Crawford}, \bibinfo{person}{Aaron Shapiro}, {and} \bibinfo{person}{Hanna
  Wallach}.} \bibinfo{year}{2017}\natexlab{}.
\newblock \bibinfo{title}{The Problem With Bias: Allocative Versus
  Representational Harms in Machine Learning}.
\newblock \bibinfo{howpublished}{Talk at the 2017 Conference of the Special
  Interest Group for Computing, Information and Society (SIGCIS), Philadelphia,
  PA}.
\newblock


\bibitem[Barocas and Selbst(2016)]%
        {barocas2016disparate}
\bibfield{author}{\bibinfo{person}{Solon Barocas} {and}
  \bibinfo{person}{Andrew~D. Selbst}.} \bibinfo{year}{2016}\natexlab{}.
\newblock \showarticletitle{Big Data's Disparate Impact}.
\newblock \bibinfo{journal}{\emph{California Law Review}}
  \bibinfo{volume}{104}, \bibinfo{number}{3} (\bibinfo{year}{2016}),
  \bibinfo{pages}{671--732}.
\newblock


\bibitem[Bertrand and Mullainathan(2004)]%
        {bertrand2004emily}
\bibfield{author}{\bibinfo{person}{Marianne Bertrand} {and}
  \bibinfo{person}{Sendhil Mullainathan}.} \bibinfo{year}{2004}\natexlab{}.
\newblock \showarticletitle{Are Emily and Greg More Employable Than Lakisha and
  Jamal? A Field Experiment on Labor Market Discrimination}.
\newblock \bibinfo{journal}{\emph{American Economic Review}}
  \bibinfo{volume}{94}, \bibinfo{number}{4} (\bibinfo{year}{2004}),
  \bibinfo{pages}{991--1013}.
\newblock
\href{https://doi.org/10.1257/0002828042002561}{doi:\nolinkurl{10.1257/0002828042002561}}


\bibitem[Black et~al\mbox{.}(2022)]%
        {black2022multiplicity}
\bibfield{author}{\bibinfo{person}{Emily Black}, \bibinfo{person}{Manish
  Raghavan}, {and} \bibinfo{person}{Solon Barocas}.}
  \bibinfo{year}{2022}\natexlab{}.
\newblock \showarticletitle{Model Multiplicity: Opportunities, Concerns, and
  Solutions}. In \bibinfo{booktitle}{\emph{Proceedings of the 2022 ACM
  Conference on Fairness, Accountability, and Transparency (FAccT)}}.
  \bibinfo{publisher}{ACM}, \bibinfo{address}{New York, NY, USA},
  \bibinfo{pages}{850--863}.
\newblock
\href{https://doi.org/10.1145/3531146.3533149}{doi:\nolinkurl{10.1145/3531146.3533149}}


\bibitem[Cheng et~al\mbox{.}(2023)]%
        {cheng2023marked}
\bibfield{author}{\bibinfo{person}{Myra Cheng}, \bibinfo{person}{Esin Durmus},
  {and} \bibinfo{person}{Dan Jurafsky}.} \bibinfo{year}{2023}\natexlab{}.
\newblock \showarticletitle{Marked Personas: Using Natural Language Prompts to
  Measure Stereotypes in Language Models}. In
  \bibinfo{booktitle}{\emph{Proceedings of the 61st Annual Meeting of the
  Association for Computational Linguistics (Volume 1: Long Papers)}}.
  \bibinfo{publisher}{Association for Computational Linguistics},
  \bibinfo{address}{Toronto, Canada}, \bibinfo{pages}{1504--1532}.
\newblock
\href{https://doi.org/10.18653/v1/2023.acl-long.84}{doi:\nolinkurl{10.18653/v1/2023.acl-long.84}}


\bibitem[{Council of the European Union}(2000a)]%
        {eu2000race}
\bibfield{author}{\bibinfo{person}{{Council of the European Union}}.}
  \bibinfo{year}{2000}\natexlab{a}.
\newblock \bibinfo{title}{Council Directive 2000/43/EC of 29 June 2000
  implementing the principle of equal treatment between persons irrespective of
  racial or ethnic origin}.
\newblock \bibinfo{howpublished}{Official Journal of the European Communities,
  OJ L 180, 19 July 2000, pp. 22--26}.
\newblock


\bibitem[{Council of the European Union}(2000b)]%
        {eu2000framework}
\bibfield{author}{\bibinfo{person}{{Council of the European Union}}.}
  \bibinfo{year}{2000}\natexlab{b}.
\newblock \bibinfo{title}{Council Directive 2000/78/EC of 27 November 2000
  establishing a general framework for equal treatment in employment and
  occupation}.
\newblock \bibinfo{howpublished}{Official Journal of the European Communities,
  OJ L 303, 2 December 2000, pp. 16--22}.
\newblock


\bibitem[Creel and Hellman(2022)]%
        {creel2022leviathan}
\bibfield{author}{\bibinfo{person}{Kathleen Creel} {and}
  \bibinfo{person}{Deborah Hellman}.} \bibinfo{year}{2022}\natexlab{}.
\newblock \showarticletitle{The Algorithmic Leviathan: Arbitrariness, Fairness,
  and Opportunity in Algorithmic Decision-Making Systems}.
\newblock \bibinfo{journal}{\emph{Canadian Journal of Philosophy}}
  \bibinfo{volume}{52}, \bibinfo{number}{1} (\bibinfo{year}{2022}),
  \bibinfo{pages}{26--43}.
\newblock
\href{https://doi.org/10.1017/can.2022.3}{doi:\nolinkurl{10.1017/can.2022.3}}


\bibitem[{CrewAI Inc.}(2024)]%
        {crewai2024}
\bibfield{author}{\bibinfo{person}{{CrewAI Inc.}}}
  \bibinfo{year}{2024}\natexlab{}.
\newblock \bibinfo{title}{{CrewAI}: Framework for orchestrating role-playing,
  autonomous AI agents}.
\newblock
\newblock
\shownote{Software. \url{https://github.com/crewAIInc/crewAI}}.


\bibitem[{DeepSeek-AI}(2025)]%
        {deepseekr1_2025}
\bibfield{author}{\bibinfo{person}{{DeepSeek-AI}}.}
  \bibinfo{year}{2025}\natexlab{}.
\newblock \bibinfo{title}{{DeepSeek-R1}: Incentivizing Reasoning Capability in
  {LLMs} via Reinforcement Learning}.
\newblock
\showeprint[arxiv]{2501.12948}


\bibitem[Deshpande et~al\mbox{.}(2023)]%
        {deshpande2023toxicity}
\bibfield{author}{\bibinfo{person}{Ameet Deshpande}, \bibinfo{person}{Vishvak
  Murahari}, \bibinfo{person}{Tanmay Rajpurohit}, \bibinfo{person}{Ashwin
  Kalyan}, {and} \bibinfo{person}{Karthik Narasimhan}.}
  \bibinfo{year}{2023}\natexlab{}.
\newblock \showarticletitle{Toxicity in {ChatGPT}: Analyzing Persona-assigned
  Language Models}. In \bibinfo{booktitle}{\emph{Findings of the Association
  for Computational Linguistics: EMNLP 2023}}. \bibinfo{publisher}{Association
  for Computational Linguistics}, \bibinfo{address}{Singapore},
  \bibinfo{pages}{1236--1270}.
\newblock
\href{https://doi.org/10.18653/v1/2023.findings-emnlp.88}{doi:\nolinkurl{10.18653/v1/2023.findings-emnlp.88}}


\bibitem[Dong et~al\mbox{.}(2024a)]%
        {dong2024not}
\bibfield{author}{\bibinfo{person}{Wenchao Dong}, \bibinfo{person}{Assem
  Zhunis}, \bibinfo{person}{Hyojin Chin}, \bibinfo{person}{Jiyoung Han}, {and}
  \bibinfo{person}{Meeyoung Cha}.} \bibinfo{year}{2024}\natexlab{a}.
\newblock \bibinfo{title}{I Am Not Them: Fluid Identities and Persistent
  Out-group Bias in Large Language Models}.
\newblock
\showeprint[arxiv]{2402.10436}~[cs.CL]


\bibitem[Dong et~al\mbox{.}(2024b)]%
        {dong2024persona}
\bibfield{author}{\bibinfo{person}{Wenchao Dong}, \bibinfo{person}{Assem
  Zhunis}, \bibinfo{person}{Dongyoung Jeong}, \bibinfo{person}{Hyojin Chin},
  \bibinfo{person}{Jiyoung Han}, {and} \bibinfo{person}{Meeyoung Cha}.}
  \bibinfo{year}{2024}\natexlab{b}.
\newblock \bibinfo{title}{Persona Setting Pitfall: Persistent Outgroup Biases
  in Large Language Models Arising from Social Identity Adoption}.
\newblock
\showeprint[arxiv]{2409.03843}~[cs.CL]


\bibitem[Dwork et~al\mbox{.}(2012)]%
        {dwork2012fairness}
\bibfield{author}{\bibinfo{person}{Cynthia Dwork}, \bibinfo{person}{Moritz
  Hardt}, \bibinfo{person}{Toniann Pitassi}, \bibinfo{person}{Omer Reingold},
  {and} \bibinfo{person}{Richard Zemel}.} \bibinfo{year}{2012}\natexlab{}.
\newblock \showarticletitle{Fairness Through Awareness}. In
  \bibinfo{booktitle}{\emph{Proceedings of the 3rd Innovations in Theoretical
  Computer Science Conference (ITCS)}}. \bibinfo{publisher}{ACM},
  \bibinfo{address}{New York, NY, USA}, \bibinfo{pages}{214--226}.
\newblock
\href{https://doi.org/10.1145/2090236.2090255}{doi:\nolinkurl{10.1145/2090236.2090255}}


\bibitem[{European Parliament and Council of the European Union}(2006)]%
        {eu2006recast}
\bibfield{author}{\bibinfo{person}{{European Parliament and Council of the
  European Union}}.} \bibinfo{year}{2006}\natexlab{}.
\newblock \bibinfo{title}{Directive 2006/54/EC of the European Parliament and
  of the Council of 5 July 2006 on the implementation of the principle of equal
  opportunities and equal treatment of men and women in matters of employment
  and occupation (recast)}.
\newblock \bibinfo{howpublished}{Official Journal of the European Union, OJ L
  204, 26 July 2006, pp. 23--36}.
\newblock


\bibitem[{European Parliament and Council of the European Union}(2024)]%
        {euaiact2024}
\bibfield{author}{\bibinfo{person}{{European Parliament and Council of the
  European Union}}.} \bibinfo{year}{2024}\natexlab{}.
\newblock \bibinfo{title}{Regulation (EU) 2024/1689 of the European Parliament
  and of the Council of 13 June 2024 laying down harmonised rules on artificial
  intelligence (Artificial Intelligence Act)}.
\newblock \bibinfo{howpublished}{Official Journal of the European Union, OJ L,
  2024/1689, 12 July 2024}.
\newblock


\bibitem[Fernandez and Fernandez-Mateo(2006)]%
        {fernandez2006networks}
\bibfield{author}{\bibinfo{person}{Roberto~M. Fernandez} {and}
  \bibinfo{person}{Isabel Fernandez-Mateo}.} \bibinfo{year}{2006}\natexlab{}.
\newblock \showarticletitle{Networks, Race, and Hiring}.
\newblock \bibinfo{journal}{\emph{American Sociological Review}}
  \bibinfo{volume}{71}, \bibinfo{number}{1} (\bibinfo{year}{2006}),
  \bibinfo{pages}{42--71}.
\newblock
\href{https://doi.org/10.1177/000312240607100103}{doi:\nolinkurl{10.1177/000312240607100103}}


\bibitem[Friedman and Nissenbaum(1996)]%
        {friedman1996bias}
\bibfield{author}{\bibinfo{person}{Batya Friedman} {and} \bibinfo{person}{Helen
  Nissenbaum}.} \bibinfo{year}{1996}\natexlab{}.
\newblock \showarticletitle{Bias in Computer Systems}.
\newblock \bibinfo{journal}{\emph{ACM Transactions on Information Systems}}
  \bibinfo{volume}{14}, \bibinfo{number}{3} (\bibinfo{year}{1996}),
  \bibinfo{pages}{330--347}.
\newblock
\href{https://doi.org/10.1145/230538.230561}{doi:\nolinkurl{10.1145/230538.230561}}


\bibitem[Gaebler et~al\mbox{.}(2024)]%
        {gaebler2024auditing}
\bibfield{author}{\bibinfo{person}{Johann~D. Gaebler}, \bibinfo{person}{Sharad
  Goel}, \bibinfo{person}{Aziz Huq}, {and} \bibinfo{person}{Prasanna Tambe}.}
  \bibinfo{year}{2024}\natexlab{}.
\newblock \bibinfo{title}{Auditing the Use of Language Models to Guide Hiring
  Decisions}.
\newblock
\showeprint[arxiv]{2404.03086}


\bibitem[Gallegos et~al\mbox{.}(2024)]%
        {gallegos2024bias}
\bibfield{author}{\bibinfo{person}{Isabel~O. Gallegos},
  \bibinfo{person}{Ryan~A. Rossi}, \bibinfo{person}{Joe Barrow},
  \bibinfo{person}{Md~Mehrab Tanjim}, \bibinfo{person}{Sungchul Kim},
  \bibinfo{person}{Franck Dernoncourt}, \bibinfo{person}{Tong Yu},
  \bibinfo{person}{Ruiyi Zhang}, {and} \bibinfo{person}{Nesreen~K. Ahmed}.}
  \bibinfo{year}{2024}\natexlab{}.
\newblock \showarticletitle{Bias and fairness in large language models: A
  survey}.
\newblock \bibinfo{journal}{\emph{Computational Linguistics}}
  \bibinfo{volume}{50}, \bibinfo{number}{3} (\bibinfo{year}{2024}),
  \bibinfo{pages}{1097--1179}.
\newblock
\href{https://doi.org/10.1162/coli_a_00524}{doi:\nolinkurl{10.1162/coli_a_00524}}


\bibitem[Geirhos et~al\mbox{.}(2020)]%
        {geirhos2020shortcut}
\bibfield{author}{\bibinfo{person}{Robert Geirhos},
  \bibinfo{person}{J{\"o}rn-Henrik Jacobsen}, \bibinfo{person}{Claudio
  Michaelis}, \bibinfo{person}{Richard Zemel}, \bibinfo{person}{Wieland
  Brendel}, \bibinfo{person}{Matthias Bethge}, {and} \bibinfo{person}{Felix~A.
  Wichmann}.} \bibinfo{year}{2020}\natexlab{}.
\newblock \showarticletitle{Shortcut Learning in Deep Neural Networks}.
\newblock \bibinfo{journal}{\emph{Nature Machine Intelligence}}
  \bibinfo{volume}{2}, \bibinfo{number}{11} (\bibinfo{year}{2020}),
  \bibinfo{pages}{665--673}.
\newblock
\href{https://doi.org/10.1038/s42256-020-00257-z}{doi:\nolinkurl{10.1038/s42256-020-00257-z}}


\bibitem[{Granite Team, IBM}(2024)]%
        {granite2024granite}
\bibfield{author}{\bibinfo{person}{{Granite Team, IBM}}.}
  \bibinfo{year}{2024}\natexlab{}.
\newblock \bibinfo{title}{{Granite 3.0} Language Models}.
\newblock
  \bibinfo{howpublished}{\url{https://github.com/ibm-granite/granite-3.0-language-models/}}.
\newblock


\bibitem[{Granite Team, IBM}(2025)]%
        {granite33_2025}
\bibfield{author}{\bibinfo{person}{{Granite Team, IBM}}.}
  \bibinfo{year}{2025}\natexlab{}.
\newblock \bibinfo{title}{{Granite-3.3-8B-Instruct}}.
\newblock
\newblock
\shownote{Model card.
  \url{https://huggingface.co/ibm-granite/granite-3.3-8b-instruct}}.


\bibitem[Greshake et~al\mbox{.}(2023)]%
        {greshake2023injection}
\bibfield{author}{\bibinfo{person}{Kai Greshake}, \bibinfo{person}{Sahar
  Abdelnabi}, \bibinfo{person}{Shailesh Mishra}, \bibinfo{person}{Christoph
  Endres}, \bibinfo{person}{Thorsten Holz}, {and} \bibinfo{person}{Mario
  Fritz}.} \bibinfo{year}{2023}\natexlab{}.
\newblock \showarticletitle{Not What You've Signed Up For: Compromising
  Real-World LLM-Integrated Applications with Indirect Prompt Injection}. In
  \bibinfo{booktitle}{\emph{Proceedings of the 16th ACM Workshop on Artificial
  Intelligence and Security (AISec)}}. \bibinfo{publisher}{ACM},
  \bibinfo{address}{Copenhagen, Denmark}, \bibinfo{pages}{79--90}.
\newblock
\href{https://doi.org/10.1145/3605764.3623985}{doi:\nolinkurl{10.1145/3605764.3623985}}


\bibitem[Gupta et~al\mbox{.}(2024)]%
        {gupta2024bias}
\bibfield{author}{\bibinfo{person}{Shashank Gupta}, \bibinfo{person}{Vaishnavi
  Shrivastava}, \bibinfo{person}{Ameet Deshpande}, \bibinfo{person}{Ashwin
  Kalyan}, \bibinfo{person}{Peter Clark}, \bibinfo{person}{Ashish Sabharwal},
  {and} \bibinfo{person}{Tushar Khot}.} \bibinfo{year}{2024}\natexlab{}.
\newblock \showarticletitle{Bias Runs Deep: Implicit Reasoning Biases in
  Persona-Assigned {LLMs}}. In \bibinfo{booktitle}{\emph{International
  Conference on Learning Representations (ICLR)}}.
  \bibinfo{publisher}{OpenReview.net}, \bibinfo{address}{Vienna, Austria},
  \bibinfo{pages}{21849--21874}.
\newblock


\bibitem[Hu et~al\mbox{.}(2025)]%
        {hu2025generative}
\bibfield{author}{\bibinfo{person}{Tiancheng Hu}, \bibinfo{person}{Yara
  Kyrychenko}, \bibinfo{person}{Steve Rathje}, \bibinfo{person}{Nigel Collier},
  \bibinfo{person}{Sander van~der Linden}, {and} \bibinfo{person}{Jon
  Roozenbeek}.} \bibinfo{year}{2025}\natexlab{}.
\newblock \showarticletitle{Generative language models exhibit social identity
  biases}.
\newblock \bibinfo{journal}{\emph{Nature Computational Science}}
  \bibinfo{volume}{5}, \bibinfo{number}{1} (\bibinfo{year}{2025}),
  \bibinfo{pages}{65--75}.
\newblock
\href{https://doi.org/10.1038/s43588-024-00741-1}{doi:\nolinkurl{10.1038/s43588-024-00741-1}}


\bibitem[Jia and Liang(2017)]%
        {jia2017adversarial}
\bibfield{author}{\bibinfo{person}{Robin Jia} {and} \bibinfo{person}{Percy
  Liang}.} \bibinfo{year}{2017}\natexlab{}.
\newblock \showarticletitle{Adversarial Examples for Evaluating Reading
  Comprehension Systems}. In \bibinfo{booktitle}{\emph{Proceedings of the 2017
  Conference on Empirical Methods in Natural Language Processing}}.
  \bibinfo{publisher}{Association for Computational Linguistics},
  \bibinfo{address}{Copenhagen, Denmark}, \bibinfo{pages}{2021--2031}.
\newblock
\href{https://doi.org/10.18653/v1/D17-1215}{doi:\nolinkurl{10.18653/v1/D17-1215}}


\bibitem[{LangChain Inc.}(2025)]%
        {langgraph2024}
\bibfield{author}{\bibinfo{person}{{LangChain Inc.}}}
  \bibinfo{year}{2025}\natexlab{}.
\newblock \bibinfo{title}{{LangGraph}: Multi-agent supervisor}.
\newblock
\newblock
\shownote{Software.
  \url{https://github.com/langchain-ai/langgraph-supervisor-py}}.


\bibitem[Lei et~al\mbox{.}(2026)]%
        {lei2026tribe}
\bibfield{author}{\bibinfo{person}{Shijun Lei}, \bibinfo{person}{Hongyu Wang},
  \bibinfo{person}{Yunji Liang}, \bibinfo{person}{Haowen Zheng},
  \bibinfo{person}{Bin Guo}, {and} \bibinfo{person}{Zhiwen Yu}.}
  \bibinfo{year}{2026}\natexlab{}.
\newblock \bibinfo{title}{Truth or Tribe: How In-group Favoritism Prioritize
  Facts in Persona Agents}.
\newblock
\showeprint[arxiv]{2605.01329}~[cs.AI]


\bibitem[Liu(2022)]%
        {llamaindex2024}
\bibfield{author}{\bibinfo{person}{Jerry Liu}.}
  \bibinfo{year}{2022}\natexlab{}.
\newblock \bibinfo{title}{{LlamaIndex}}.
\newblock
\newblock
\shownote{Software. \url{https://github.com/run-llama/llama_index}}.


\bibitem[MacAvaney et~al\mbox{.}(2022)]%
        {macavaney2022abnirml}
\bibfield{author}{\bibinfo{person}{Sean MacAvaney}, \bibinfo{person}{Sergey
  Feldman}, \bibinfo{person}{Nazli Goharian}, \bibinfo{person}{Doug Downey},
  {and} \bibinfo{person}{Arman Cohan}.} \bibinfo{year}{2022}\natexlab{}.
\newblock \showarticletitle{{ABNIRML}: Analyzing the Behavior of Neural {IR}
  Models}.
\newblock \bibinfo{journal}{\emph{Transactions of the Association for
  Computational Linguistics}}  \bibinfo{volume}{10} (\bibinfo{year}{2022}),
  \bibinfo{pages}{224--239}.
\newblock
\href{https://doi.org/10.1162/tacl_a_00457}{doi:\nolinkurl{10.1162/tacl_a_00457}}


\bibitem[McCoy et~al\mbox{.}(2019)]%
        {mccoy2019right}
\bibfield{author}{\bibinfo{person}{Tom McCoy}, \bibinfo{person}{Ellie Pavlick},
  {and} \bibinfo{person}{Tal Linzen}.} \bibinfo{year}{2019}\natexlab{}.
\newblock \showarticletitle{Right for the Wrong Reasons: Diagnosing Syntactic
  Heuristics in Natural Language Inference}. In
  \bibinfo{booktitle}{\emph{Proceedings of the 57th Annual Meeting of the
  Association for Computational Linguistics}}. \bibinfo{publisher}{Association
  for Computational Linguistics}, \bibinfo{address}{Florence, Italy},
  \bibinfo{pages}{3428--3448}.
\newblock
\href{https://doi.org/10.18653/v1/P19-1334}{doi:\nolinkurl{10.18653/v1/P19-1334}}


\bibitem[Metaxa et~al\mbox{.}(2021)]%
        {metaxa2021auditing}
\bibfield{author}{\bibinfo{person}{Danaë Metaxa}, \bibinfo{person}{Joon~Sung
  Park}, \bibinfo{person}{Ronald~E. Robertson}, \bibinfo{person}{Karrie
  Karahalios}, \bibinfo{person}{Christo Wilson}, \bibinfo{person}{Jeff
  Hancock}, {and} \bibinfo{person}{Christian Sandvig}.}
  \bibinfo{year}{2021}\natexlab{}.
\newblock \showarticletitle{Auditing Algorithms: Understanding Algorithmic
  Systems from the Outside In}.
\newblock \bibinfo{journal}{\emph{Foundations and Trends in Human-Computer
  Interaction}} \bibinfo{volume}{14}, \bibinfo{number}{4}
  (\bibinfo{year}{2021}), \bibinfo{pages}{272--344}.
\newblock
\href{https://doi.org/10.1561/1100000083}{doi:\nolinkurl{10.1561/1100000083}}


\bibitem[{Microsoft}(2024)]%
        {autogen2024software}
\bibfield{author}{\bibinfo{person}{{Microsoft}}.}
  \bibinfo{year}{2024}\natexlab{}.
\newblock \bibinfo{title}{{AutoGen}: A programming framework for agentic {AI}}.
\newblock
\newblock
\shownote{Software. \url{https://github.com/microsoft/autogen}}.


\bibitem[Moon et~al\mbox{.}(2026)]%
        {moon2026identity}
\bibfield{author}{\bibinfo{person}{Suhong Moon}, \bibinfo{person}{Minwoo Kang},
  \bibinfo{person}{Joseph Suh}, \bibinfo{person}{Mustafa Safdari}, {and}
  \bibinfo{person}{John Canny}.} \bibinfo{year}{2026}\natexlab{}.
\newblock \bibinfo{title}{Identity, Cooperation and Framing Effects within
  Groups of Real and Simulated Humans}.
\newblock
\showeprint[arxiv]{2601.16355}~[cs.CL]


\bibitem[{New York City Council}(2021)]%
        {nyc2021ll144}
\bibfield{author}{\bibinfo{person}{{New York City Council}}.}
  \bibinfo{year}{2021}\natexlab{}.
\newblock \bibinfo{title}{Local Law 144 of 2021: Automated Employment Decision
  Tools}.
\newblock \bibinfo{howpublished}{N.Y.C. Admin. Code \S\S 20-870 to 20-874;
  final rules adopted by the Department of Consumer and Worker Protection,
  effective 5 July 2023}.
\newblock


\bibitem[Obermeyer et~al\mbox{.}(2019)]%
        {obermeyer2019dissecting}
\bibfield{author}{\bibinfo{person}{Ziad Obermeyer}, \bibinfo{person}{Brian
  Powers}, \bibinfo{person}{Christine Vogeli}, {and} \bibinfo{person}{Sendhil
  Mullainathan}.} \bibinfo{year}{2019}\natexlab{}.
\newblock \showarticletitle{Dissecting Racial Bias in an Algorithm Used to
  Manage the Health of Populations}.
\newblock \bibinfo{journal}{\emph{Science}} \bibinfo{volume}{366},
  \bibinfo{number}{6464} (\bibinfo{year}{2019}), \bibinfo{pages}{447--453}.
\newblock
\href{https://doi.org/10.1126/science.aax2342}{doi:\nolinkurl{10.1126/science.aax2342}}


\bibitem[Olsson et~al\mbox{.}(2022)]%
        {olsson2022induction}
\bibfield{author}{\bibinfo{person}{Catherine Olsson}, \bibinfo{person}{Nelson
  Elhage}, \bibinfo{person}{Neel Nanda}, \bibinfo{person}{Nicholas Joseph},
  \bibinfo{person}{Nova DasSarma}, \bibinfo{person}{Tom Henighan},
  \bibinfo{person}{Ben Mann}, \bibinfo{person}{Amanda Askell},
  \bibinfo{person}{Yuntao Bai}, \bibinfo{person}{Anna Chen},
  \bibinfo{person}{Tom Conerly}, \bibinfo{person}{Dawn Drain},
  \bibinfo{person}{Deep Ganguli}, \bibinfo{person}{Zac Hatfield-Dodds},
  \bibinfo{person}{Danny Hernandez}, \bibinfo{person}{Scott Johnston},
  \bibinfo{person}{Andy Jones}, \bibinfo{person}{Jackson Kernion},
  \bibinfo{person}{Liane Lovitt}, \bibinfo{person}{Kamal Ndousse},
  \bibinfo{person}{Dario Amodei}, \bibinfo{person}{Tom Brown},
  \bibinfo{person}{Jack Clark}, \bibinfo{person}{Jared Kaplan},
  \bibinfo{person}{Sam McCandlish}, {and} \bibinfo{person}{Chris Olah}.}
  \bibinfo{year}{2022}\natexlab{}.
\newblock \bibinfo{title}{In-context Learning and Induction Heads}.
\newblock \bibinfo{howpublished}{Transformer Circuits Thread}.
\newblock
\newblock
\shownote{\url{https://transformer-circuits.pub/2022/in-context-learning-and-induction-heads/}}.


\bibitem[{OpenAI}(2025)]%
        {openaiagents2025}
\bibfield{author}{\bibinfo{person}{{OpenAI}}.} \bibinfo{year}{2025}\natexlab{}.
\newblock \bibinfo{title}{{OpenAI Agents SDK}}.
\newblock
\newblock
\shownote{Software. \url{https://github.com/openai/openai-agents-python}}.


\bibitem[Panickssery et~al\mbox{.}(2024)]%
        {panickssery2024llm}
\bibfield{author}{\bibinfo{person}{Arjun Panickssery},
  \bibinfo{person}{Samuel~R. Bowman}, {and} \bibinfo{person}{Shi Feng}.}
  \bibinfo{year}{2024}\natexlab{}.
\newblock \showarticletitle{{LLM} Evaluators Recognize and Favor Their Own
  Generations}. In \bibinfo{booktitle}{\emph{Advances in Neural Information
  Processing Systems}}, Vol.~\bibinfo{volume}{37}. \bibinfo{publisher}{Neural
  Information Processing Systems Foundation}, \bibinfo{address}{Vancouver, BC,
  Canada}, \bibinfo{pages}{68772--68802}.
\newblock
\href{https://doi.org/10.52202/079017-2197}{doi:\nolinkurl{10.52202/079017-2197}}


\bibitem[Prince and Schwarcz(2020)]%
        {prince2020proxy}
\bibfield{author}{\bibinfo{person}{Anya E.~R. Prince} {and}
  \bibinfo{person}{Daniel Schwarcz}.} \bibinfo{year}{2020}\natexlab{}.
\newblock \showarticletitle{Proxy Discrimination in the Age of Artificial
  Intelligence and Big Data}.
\newblock \bibinfo{journal}{\emph{Iowa Law Review}} \bibinfo{volume}{105},
  \bibinfo{number}{3} (\bibinfo{year}{2020}), \bibinfo{pages}{1257--1318}.
\newblock


\bibitem[Raghavan et~al\mbox{.}(2020)]%
        {raghavan2020mitigating}
\bibfield{author}{\bibinfo{person}{Manish Raghavan}, \bibinfo{person}{Solon
  Barocas}, \bibinfo{person}{Jon Kleinberg}, {and} \bibinfo{person}{Karen
  Levy}.} \bibinfo{year}{2020}\natexlab{}.
\newblock \showarticletitle{Mitigating Bias in Algorithmic Hiring: Evaluating
  Claims and Practices}. In \bibinfo{booktitle}{\emph{Proceedings of the 2020
  Conference on Fairness, Accountability, and Transparency (FAT*)}}.
  \bibinfo{publisher}{ACM}, \bibinfo{address}{Barcelona, Spain},
  \bibinfo{pages}{469--481}.
\newblock
\href{https://doi.org/10.1145/3351095.3372828}{doi:\nolinkurl{10.1145/3351095.3372828}}


\bibitem[Raji and Buolamwini(2019)]%
        {raji2019actionable}
\bibfield{author}{\bibinfo{person}{Inioluwa~Deborah Raji} {and}
  \bibinfo{person}{Joy Buolamwini}.} \bibinfo{year}{2019}\natexlab{}.
\newblock \showarticletitle{Actionable Auditing: Investigating the Impact of
  Publicly Naming Biased Performance Results of Commercial {AI} Products}. In
  \bibinfo{booktitle}{\emph{Proceedings of the 2019 AAAI/ACM Conference on AI,
  Ethics, and Society (AIES)}}. \bibinfo{publisher}{ACM}, \bibinfo{address}{New
  York, NY, USA}, \bibinfo{pages}{429--435}.
\newblock
\href{https://doi.org/10.1145/3306618.3314244}{doi:\nolinkurl{10.1145/3306618.3314244}}


\bibitem[Sclar et~al\mbox{.}(2024)]%
        {sclar2024formatspread}
\bibfield{author}{\bibinfo{person}{Melanie Sclar}, \bibinfo{person}{Yejin
  Choi}, \bibinfo{person}{Yulia Tsvetkov}, {and} \bibinfo{person}{Alane Suhr}.}
  \bibinfo{year}{2024}\natexlab{}.
\newblock \showarticletitle{Quantifying Language Models' Sensitivity to
  Spurious Features in Prompt Design or: How I Learned to Start Worrying About
  Prompt Formatting}. In \bibinfo{booktitle}{\emph{International Conference on
  Learning Representations (ICLR)}}. \bibinfo{publisher}{OpenReview.net},
  \bibinfo{address}{Vienna, Austria}, \bibinfo{pages}{25055--25083}.
\newblock


\bibitem[Seshadri et~al\mbox{.}(2025)]%
        {seshadri2025small}
\bibfield{author}{\bibinfo{person}{Preethi Seshadri}, \bibinfo{person}{Hongyu
  Chen}, \bibinfo{person}{Sameer Singh}, {and} \bibinfo{person}{Seraphina
  Goldfarb-Tarrant}.} \bibinfo{year}{2025}\natexlab{}.
\newblock \showarticletitle{Small Changes, Large Consequences: Analyzing the
  Allocational Fairness of {LLMs} in Hiring Contexts}. In
  \bibinfo{booktitle}{\emph{Proceedings of the 14th International Joint
  Conference on Natural Language Processing and the 4th Conference of the
  Asia-Pacific Chapter of the Association for Computational Linguistics}}.
  \bibinfo{publisher}{Association for Computational Linguistics},
  \bibinfo{address}{Mumbai, India}, \bibinfo{pages}{2645--2665}.
\newblock
\href{https://doi.org/10.18653/v1/2025.ijcnlp-long.143}{doi:\nolinkurl{10.18653/v1/2025.ijcnlp-long.143}}


\bibitem[Shi et~al\mbox{.}(2023)]%
        {shi2023distracted}
\bibfield{author}{\bibinfo{person}{Freda Shi}, \bibinfo{person}{Xinyun Chen},
  \bibinfo{person}{Kanishka Misra}, \bibinfo{person}{Nathan Scales},
  \bibinfo{person}{David Dohan}, \bibinfo{person}{Ed~H. Chi},
  \bibinfo{person}{Nathanael Sch{\"a}rli}, {and} \bibinfo{person}{Denny Zhou}.}
  \bibinfo{year}{2023}\natexlab{}.
\newblock \showarticletitle{Large Language Models Can Be Easily Distracted by
  Irrelevant Context}. In \bibinfo{booktitle}{\emph{Proceedings of the 40th
  International Conference on Machine Learning}}
  \emph{(\bibinfo{series}{Proceedings of Machine Learning Research},
  Vol.~\bibinfo{volume}{202})}. \bibinfo{publisher}{PMLR},
  \bibinfo{address}{Honolulu, HI, USA}, \bibinfo{pages}{31210--31227}.
\newblock


\bibitem[Tamkin et~al\mbox{.}(2023)]%
        {tamkin2023evaluating}
\bibfield{author}{\bibinfo{person}{Alex Tamkin}, \bibinfo{person}{Amanda
  Askell}, \bibinfo{person}{Liane Lovitt}, \bibinfo{person}{Esin Durmus},
  \bibinfo{person}{Nicholas Joseph}, \bibinfo{person}{Shauna Kravec},
  \bibinfo{person}{Karina Nguyen}, \bibinfo{person}{Jared Kaplan}, {and}
  \bibinfo{person}{Deep Ganguli}.} \bibinfo{year}{2023}\natexlab{}.
\newblock \bibinfo{title}{Evaluating and Mitigating Discrimination in Language
  Model Decisions}.
\newblock
\showeprint[arxiv]{2312.03689}~[cs.CL]


\bibitem[{United States Congress}(1991)]%
        {titlevii1991}
\bibfield{author}{\bibinfo{person}{{United States Congress}}.}
  \bibinfo{year}{1991}\natexlab{}.
\newblock \bibinfo{title}{Civil Rights Act of 1964, Title VII, as amended by
  the Civil Rights Act of 1991}.
\newblock \bibinfo{howpublished}{42 U.S.C. \S~2000e-2(k)}.
\newblock


\bibitem[Wang et~al\mbox{.}(2024)]%
        {wang2024fairevaluators}
\bibfield{author}{\bibinfo{person}{Peiyi Wang}, \bibinfo{person}{Lei Li},
  \bibinfo{person}{Liang Chen}, \bibinfo{person}{Zefan Cai},
  \bibinfo{person}{Dawei Zhu}, \bibinfo{person}{Binghuai Lin},
  \bibinfo{person}{Yunbo Cao}, \bibinfo{person}{Lingpeng Kong},
  \bibinfo{person}{Qi Liu}, \bibinfo{person}{Tianyu Liu}, {and}
  \bibinfo{person}{Zhifang Sui}.} \bibinfo{year}{2024}\natexlab{}.
\newblock \showarticletitle{Large Language Models are not Fair Evaluators}. In
  \bibinfo{booktitle}{\emph{Proceedings of the 62nd Annual Meeting of the
  Association for Computational Linguistics (Volume 1: Long Papers)}}.
  \bibinfo{publisher}{Association for Computational Linguistics},
  \bibinfo{address}{Bangkok, Thailand}, \bibinfo{pages}{9440--9450}.
\newblock
\href{https://doi.org/10.18653/v1/2024.acl-long.511}{doi:\nolinkurl{10.18653/v1/2024.acl-long.511}}


\bibitem[Wang et~al\mbox{.}(2026)]%
        {wang2026outgroup}
\bibfield{author}{\bibinfo{person}{Zongwei Wang}, \bibinfo{person}{Bincheng
  Gu}, \bibinfo{person}{Hongyu Yu}, \bibinfo{person}{Junliang Yu},
  \bibinfo{person}{Tao He}, \bibinfo{person}{Jiayin Feng},
  \bibinfo{person}{Chenghua Lin}, {and} \bibinfo{person}{Min Gao}.}
  \bibinfo{year}{2026}\natexlab{}.
\newblock \bibinfo{title}{When Agents See Humans as the Outgroup:
  Belief-Dependent Bias in LLM-Powered Agents}.
\newblock
\showeprint[arxiv]{2601.00240}~[cs.AI]


\bibitem[Wilson and Caliskan(2024)]%
        {wilson2024gender}
\bibfield{author}{\bibinfo{person}{Kyra Wilson} {and} \bibinfo{person}{Aylin
  Caliskan}.} \bibinfo{year}{2024}\natexlab{}.
\newblock \showarticletitle{Gender, Race, and Intersectional Bias in Resume
  Screening via Language Model Retrieval}. In
  \bibinfo{booktitle}{\emph{Proceedings of the AAAI/ACM Conference on AI,
  Ethics, and Society (AIES)}}, Vol.~\bibinfo{volume}{7}.
  \bibinfo{publisher}{Association for the Advancement of Artificial
  Intelligence}, \bibinfo{address}{San Jose, CA, USA},
  \bibinfo{pages}{1578--1590}.
\newblock
\href{https://doi.org/10.1609/aies.v7i1.31748}{doi:\nolinkurl{10.1609/aies.v7i1.31748}}


\bibitem[Wright et~al\mbox{.}(2024)]%
        {wright2024null}
\bibfield{author}{\bibinfo{person}{Lucas Wright}, \bibinfo{person}{Roxana~Mika
  Muenster}, \bibinfo{person}{Briana Vecchione}, \bibinfo{person}{Tianyao Qu},
  \bibinfo{person}{Pika~(Senhuang) Cai}, \bibinfo{person}{Alan Smith},
  \bibinfo{person}{{COMM/INFO 2450 Student Investigators}},
  \bibinfo{person}{Jacob Metcalf}, {and} \bibinfo{person}{J.~Nathan Matias}.}
  \bibinfo{year}{2024}\natexlab{}.
\newblock \showarticletitle{Null Compliance: {NYC} Local Law 144 and the
  Challenges of Algorithm Accountability}. In
  \bibinfo{booktitle}{\emph{Proceedings of the 2024 ACM Conference on Fairness,
  Accountability, and Transparency (FAccT)}}. \bibinfo{publisher}{ACM},
  \bibinfo{address}{New York, NY, USA}, \bibinfo{pages}{1701--1713}.
\newblock
\href{https://doi.org/10.1145/3630106.3658998}{doi:\nolinkurl{10.1145/3630106.3658998}}


\bibitem[Wu et~al\mbox{.}(2023)]%
        {wu2023autogen}
\bibfield{author}{\bibinfo{person}{Qingyun Wu}, \bibinfo{person}{Gagan Bansal},
  \bibinfo{person}{Jieyu Zhang}, \bibinfo{person}{Yiran Wu},
  \bibinfo{person}{Beibin Li}, \bibinfo{person}{Erkang Zhu},
  \bibinfo{person}{Li Jiang}, \bibinfo{person}{Xiaoyun Zhang},
  \bibinfo{person}{Shaokun Zhang}, \bibinfo{person}{Jiale Liu},
  \bibinfo{person}{Ahmed~Hassan Awadallah}, \bibinfo{person}{Ryen~W White},
  \bibinfo{person}{Doug Burger}, {and} \bibinfo{person}{Chi Wang}.}
  \bibinfo{year}{2023}\natexlab{}.
\newblock \bibinfo{title}{{AutoGen}: Enabling Next-Gen {LLM} Applications via
  Multi-Agent Conversation}.
\newblock
\showeprint[arxiv]{2308.08155}


\bibitem[Yang et~al\mbox{.}(2025)]%
        {qwen3technical2025}
\bibfield{author}{\bibinfo{person}{An Yang} {et~al\mbox{.}}}
  \bibinfo{year}{2025}\natexlab{}.
\newblock \bibinfo{title}{Qwen3 Technical Report}.
\newblock
\showeprint[arxiv]{2505.09388}~[cs.CL]


\bibitem[Zheng et~al\mbox{.}(2024)]%
        {zheng2024selectors}
\bibfield{author}{\bibinfo{person}{Chujie Zheng}, \bibinfo{person}{Hao Zhou},
  \bibinfo{person}{Fandong Meng}, \bibinfo{person}{Jie Zhou}, {and}
  \bibinfo{person}{Minlie Huang}.} \bibinfo{year}{2024}\natexlab{}.
\newblock \showarticletitle{Large Language Models Are Not Robust Multiple
  Choice Selectors}. In \bibinfo{booktitle}{\emph{International Conference on
  Learning Representations (ICLR)}}. \bibinfo{publisher}{OpenReview.net},
  \bibinfo{address}{Vienna, Austria}, \bibinfo{pages}{19426--19454}.
\newblock


\end{thebibliography}

\appendix

\section{Instrument and method detail}

\subsection{Checks on the instrument}
\label{sec:checks}

\begin{table}[h]
\centering
\caption{The fourteen models. Six families contribute a base and an
instruction-tuned model, and two contribute an instruction-tuned model only.
We count the reasoning model among the instruction-tuned models. The Qwen3
instruction-tuned model is the hybrid checkpoint \texttt{Qwen3-8B}, and it and
Granite3.3 were run with their reasoning mode switched off.}
\label{tab:models}
\small
\begin{tabular}{@{}llll@{}}
\toprule
Family & Size & Tuned & Base \\
\midrule
Falcon3                                   & 10B & Instruct  & yes \\
Gemma2                                    & 9B  & Instruct  & yes \\
Llama3.1                                  & 8B  & Instruct  & yes \\
Mistral-Nemo                              & 12B & Instruct  & yes \\
OLMo2                                     & 7B  & Instruct  & yes \\
Qwen3 \citep{qwen3technical2025}           & 8B  & Instruct  & yes \\
Granite3.3 \citep{granite2024granite, granite33_2025}      & 8B  & Instruct  & no  \\
DeepSeek-R1-Distill \citep{deepseekr1_2025} & 14B & Reasoning & no  \\
\bottomrule
\end{tabular}
\end{table}

Median winning-letter log-probabilities ranged from $-1.822$ to $-0.000$ across
cells, and no model placed any trial below the $-8.0$ threshold. Two
further models, Phi-4-reasoning and SOLAR-10.7B-Instruct, were excluded
before collection on the basis of the same check applied to the companion
study's data. On this study's own resource pilots, Phi-4-reasoning also
failed (median $-19.1$, every trial below $-8$), but SOLAR passed (median
$-3.6$, 9.2\% of trials below $-8$). Its exclusion therefore rests on the
companion data alone. The pilots (5{,}400 trials, including a Study 4 pilot
on one model) only sized the collection and are not analysed. The two later passes were gated the same way and
passed independently, at $-1.685$ to $-0.000$ and $-1.548$ to $-0.000$, with no
trial below threshold and criterion responsiveness of 88.4\% and 89.3\% of
cells against 88.6\% overall.

The pooled responsiveness figure is set almost entirely by the base
models. Instruction-tuned cells respond to the competing criterion in
99.7\% of cases and base cells in 73.8\%, and two base models,
OLMo2-7B and Llama3.1-8B, account for 141 of the 153 non-responding cells.
Every headline in this paper is computed on the instruction-tuned models,
where the check effectively never fails.

Dropping the non-responding cells rather than reporting them raises every
effect and leaves the no-effect conditions where they are. The \texttt{match} median goes from
$+0.352$ to $+0.490$, \texttt{conflict} from $+0.266$ to $+0.374$,
\texttt{nonpersona} from $+0.178$ to $+0.265$ and \texttt{nearmiss} from
$+0.045$ to $+0.096$, while \texttt{neutral} stays at $+0.002$. The
instruction-tuned exchange rates are unchanged to two decimal places, because
those cells all respond. We report the full set, which is the conservative
choice.

The estimator in Section~\ref{sec:estimating} fits the paired contrast as
linear in the merit level index. Adding a quadratic term in that index to each
of the 168 \texttt{match} cells gains a median 0.3\% of the variance, against a
median $R^2$ of 0.84 for the linear fit in instruction-tuned cells. The linear
form is adequate for these data.

Model weights are bf16, so the mantissa spacing places each per-trial $\delta$
on a lattice. The spacing follows the model's logit magnitude rather than
being one constant, running from $0.0625$ to $0.25$ in the eight models
where a single spacing fits at least 95\% of trials. In the other six no
single spacing down to $0.03125$ fits.
This quantisation adds noise and shrinks the estimate. It does not create
structure, and the control conditions sit at zero under it
(Table~\ref{tab:arms}). The discrete readout is unaffected, since it depends
only on the sign of $\delta$.

\begin{table*}[t]
\centering
\caption{The eight conditions. In \texttt{match}, \texttt{neutral},
\texttt{nonpersona}, \texttt{conflict} and \texttt{nearmiss} both records carry
a real attribute and the manipulation is which one the prompt shares.
\texttt{blind} and \texttt{blind\_filler} show no attribute.
\texttt{conflict\_control} varies the field value alone, holding field count,
names and order identical across the two records.}
\label{tab:arms-design}
\begin{tabular}{lp{0.62\linewidth}}
\toprule
Condition & Construction \\
\midrule
\texttt{match} & Persona carries the attribute of one candidate. \\
\texttt{neutral} & Persona carries an attribute matching neither candidate, separating a match from the mere presence of a persona. \\
\texttt{blind} & Attribute absent from the records entirely, and the persona names no string. \\
\texttt{blind\_filler} & Attribute replaced by a constant field identical on both records, holding record length fixed. The persona names no string. \\
\texttt{nonpersona} & Persona names no string; one candidate's string appears in a settings field beneath the persona. \\
\midrule
\texttt{conflict} & Persona exactly as in \texttt{match} and the record strings byte-identical, but the holding field is renamed \texttt{conflict\_declared}. \\
\texttt{conflict\_control} & The field without the shared string: persona neutral, one record declares a conflict and the other declares \texttt{none}. Tracks the \emph{declaring} candidate. \\
\texttt{nearmiss} & Persona and one record carry sibling strings sharing every distinguishing token but denoting different units: \texttt{project-kestrel-alpha} against \texttt{project-kestrel-beta}, \texttt{Osaka North} against \texttt{Osaka South}, \texttt{Ashford University} against \texttt{Ashford College}. \\
\bottomrule
\end{tabular}
\end{table*}

\subsection{Construction of the mechanism conditions}
\label{sec:armconstruction}

Two construction choices in the mechanism conditions carry the inference. In \texttt{conflict} the field value is byte-identical to \texttt{match}, so
the comparison is not confounded by the string. Nothing in the prompt
instructs the model to weigh the field. The norm has to come from the model. In
\texttt{nearmiss} under \texttt{v\_inert} and \texttt{v\_org}, the
distinguishing qualifier is applied to both sides deliberately. A bare \texttt{project-kestrel} against
\texttt{project-kestrel-beta} would read as a parent and its own sub-project.
That is a scope relation, and it would defeat the condition. A near-miss sharing only
the vocabulary's common head would be empty by construction, because under
\texttt{v\_inert} the \texttt{neutral} attribute already shares that head.

\subsection{Reading the binomial intervals}
\label{sec:intervals}

Intervals on discrete rates are normal-approximation binomial intervals
($\pm1.96$ standard errors) on the trials pooled for that rate.
Two cautions apply to them. The two orders of one pair are not
independent. A pooled rate also averages models that differ widely, and a
binomial interval describes sampling within that pool rather than the spread
between models. We therefore read these intervals as an indication of
precision and not as a test, and we report the between-model spread
separately wherever we pool.

\subsection{The family-clustered sign test}
\label{sec:familytest}

The exact sign test assumes the models are independent, and ours are not fully
independent. Six families contribute a base and a tuned model sharing
weights. We therefore run the same test a second time clustered by family, with
each of the eight families contributing the median of its own models. We
report both counts and read the family-level one as the conservative figure.
Table~\ref{tab:arms} gives both. The $p$-values we report are unadjusted.
The four main-text family tests at $p = .008$ remain below $.05$ after a Holm
correction over those four ($p = .031$). The 14-model tests survive any
correction. Across all eight conditions of Table~\ref{tab:arms}, a
Bonferroni correction would raise $.008$ to $.063$.

\subsection{Collection passes}
\label{sec:passes}

The mechanism conditions were collected in two later passes, after the first five had
been analysed, and no earlier condition was recollected. The generator draws each
condition's records from its own named stream, so a condition's records do not depend on
which other conditions were requested. The three passes share a design, pools,
domains, merit grid and both candidate orders, but not a random stream.
Appendix~\ref{sec:repro} records why, and reports the checks on the streams.

\subsection{Position bias, and presenting each pair in both orders}
\label{sec:positionprior}

These models carry a position bias large enough to dominate the discrete choice
outright; in one model the first option receives a log-probability of
$-0.00$ against $-8.50$ on a matched pair. We therefore present each pair
twice, in both orders. On the continuous readout the bias cancels when the two
orders are differenced. The cancellation is close rather than exact, because
the second order redraws two settings-block values, \texttt{assistant\_id}
and \texttt{policy\_version}. On the discrete readout we average the indicator ``the
selected letter is the sharing candidate's'' over both orders, which cancels
the bias on average rather than by construction. The
\texttt{neutral} condition returns 0.501 at equal merit against an expected 0.500,
which is the check that licenses reading the discrete rates as effects on
decisions. Intervals on discrete rates are binomial on the trials in the cell; Appendix~\ref{sec:intervals} says how to read them.

\begin{table*}[t]
\centering
\caption{Allocation domains, their competing criteria, and the five levels the
grid steps through. The grid manipulates the level \emph{index}, so a merit
difference of one is one step along the row and not one native unit.}
\label{tab:domains}
\begin{tabular}{llll}
\toprule
Domain & Decision & Competing criterion & Levels \\
\midrule
\texttt{shortlist} & Advance one applicant & Years of relevant experience & 1, 3, 5, 8, 12 \\
\texttt{claims} & Approve one claim & Supporting documents filed & 1, 2, 4, 6, 9 \\
\texttt{referral} & Prioritise one referral & Clinical urgency score & 2, 4, 6, 8, 10 \\
\texttt{scholarship} & Award to one student & Financial need index & 20, 40, 60, 75, 90 \\
\bottomrule
\end{tabular}
\end{table*}

\subsection{On the word merit}
\label{sec:meritword}

We call this criterion \emph{merit} throughout, and the word needs one
qualification. It names whatever the scenario stipulates as the ground of the
decision. In \texttt{scholarship} that is financial need, which is not
something a candidate earned. We take no position on whether any of the four
criteria is the right basis for the decision it governs. The design requires
only that the criterion be the one the scenario puts in front of the model. A
shift toward the sharing candidate is then a shift away from what the decision
is supposed to turn on, whatever one thinks of that basis.

\subsection{Persona vocabularies in detail}
\label{sec:vocabnotes}

\texttt{v\_inert} comprises made-up project codes, rendered as ``You work
on \texttt{project-kestrel}'' and shared with a \texttt{req\_ref} field. \texttt{v\_org} comprises city
names rendered as ``You are based in the Osaka office'' and shared with an
\texttt{office} field. \texttt{v\_institution} comprises university
names rendered as ``You are a graduate of Ashford University'' and shared with
an \texttt{institution} field. The three differ in the social content they
carry. They also differ in how easily the shared string reads as naming the
model's own job. Section~\ref{sec:mechanism} takes up that second
difference. A fourth vocabulary, \texttt{v\_race}, carries the candidate's race as an
explicit field. Its levels are not exchangeable with one another, so it has its
own design and estimator; Section~\ref{sec:race} gives both.

\texttt{v\_org} is the only vocabulary made of common real place names. Its
cells therefore differ from the other two in how often the words occur in
text and in how the model splits them into pieces, as well as in meaning. We flag this where it bears on results.

\subsection{Study 4 estimator}
\label{sec:raceestimator}

Study 1 fits the paired contrast as $\beta_{\text{match}} + m\,\beta_{\text{merit}}$.
Study 4 adds one term:
\begin{equation}
\text{contrast} = \beta_{\text{match}} + m\,\beta_{\text{merit}}
  + \left(\gamma_{r_t} - \gamma_{r_o}\right),
\end{equation}
where $r_t$ is the race of the sharing candidate and $r_o$ the race of the
other. $\beta_{\text{match}}$ is the matching effect, averaged over which race
the persona carries. $\gamma_r$ is how much more often race $r$ is selected
whatever the persona says, with White as the reference and pinned at zero.

We summarise the size of the fixed ordering by its spread, the largest
$\gamma_r$ minus the smallest within one model-by-domain cell, taking the
median over the 32 cells of a condition. The spread was 0.77 in
\texttt{neutral}, which is the cleaner estimate because no race is shared
there. It was 1.45 in \texttt{match}, where part of the larger shared-string
effect for Black candidates than for White ones (Table~\ref{tab:race-by-race})
is absorbed into $\gamma$, so the additive fit does not hold exactly. As a
check, the design assigns a race to every record in \texttt{blind} and
\texttt{blind\_filler} as well, but the model never sees it. Fitting those
assigned races returns a spread of 0.07, which is the value the estimator
gives when no race reaches the model. White was the least-selected race on
both readouts. Black and Hispanic were the most selected. In \texttt{neutral}
Black is above Hispanic on both readouts, and in \texttt{match} the two swap
places between the log-odds and the decisions. The White-candidate effect in
\texttt{scholarship} was $+0.03$ in decision rate, indistinguishable from
zero.

\subsection{The merit exchange rate}
\label{sec:exchangedetail}

Within a cell we fit the log-odds $\delta = \log p(A) - \log p(B)$ by
least squares. Let $s = +1$ when the sharing candidate is option A and
$s = -1$ when it is option B. Let $m$ be the sharing candidate's merit
advantage, in levels, so that $m = 0$ means the two candidates are equally
qualified and $m = -1$ means the sharing candidate is one level weaker. We fit

\begin{equation}
\delta = \alpha + s\,\beta_{\text{match}} + s\,m\,\beta_{\text{merit}} + \epsilon,
\end{equation}

\noindent where $\alpha$ is the position bias. Because every pair appears in
both orders, half the difference between the two orders equals
$\beta_{\text{match}} + m\,\beta_{\text{merit}}$, and the position bias
cancels. $\beta_{\text{merit}}$ is the effect of one merit level.
$\beta_{\text{match}}$ is the lean toward the sharing candidate at equal
merit. In a condition with no shared string, $s$ marks the record that would
share the string under persona match. The lean is close to linear in merit
level: adding a curved term explains a median 0.3\% more of the variation
(Appendix~\ref{sec:checks}). Decision rates flatten at the extremes, as in
Table~\ref{tab:discrete}, because a lean beyond a certain size always
produces the same decision.

The ratio $\beta_{\text{match}} / \beta_{\text{merit}}$ is the \emph{merit exchange rate}. It is the extra merit a
candidate without the shared string needs to be selected as often. It is a
ratio of estimates, so a cluster bootstrap over replicates gives its interval
within a cell. Across cells we report the median and the interquartile range
instead, because the spread there is between models rather than sampling
error. The rate is expressed in \emph{merit levels}, because $m$ is a level
index. The level spacings in Table~\ref{tab:domains} are deliberately non-uniform. A
coefficient on the level index therefore has no single equivalent in years or
documents. We convert to a native unit only once, in
Section~\ref{sec:exchange}, with the spacing stated.

\subsection{Deriving the validity threshold}

For a model whose chat template auto-opens a reasoning block, the answer letters
sit far into the tail, where differences are numerical noise on vanishing
probability mass. A model is therefore excluded before analysis if its
median winning-letter log-probability falls below $-8.0$, or if it places more
than 10\% of trials below that value.

\subsection{One trial, verbatim}
\label{sec:exampleprompt}

Figure~\ref{fig:prompt} gives the exact text of one \texttt{match} trial in
\texttt{shortlist} under \texttt{v\_inert}, at equal merit. The persona
occupies the system prompt and the two records occupy the user turn. Nothing
in either turn says the \texttt{req\_ref} field is relevant. The filler fields
are drawn identically for the two records, so the records differ only in
\texttt{applicant\_id} and in the field carrying the shared string. The paired
trial exchanges the two records and redraws \texttt{assistant\_id} and
\texttt{policy\_version}. For Gemma2 and Granite3.3 the system text opens the
user turn instead (Section~\ref{sec:readout}).

\begin{figure}[t]
\scriptsize
\begin{verbatim}
[system]
You are a shortlisting assistant for the Merrow
recruiting team. You review applications and advance
the stronger candidate for interview. You work on
project-kestrel.
Instance metadata:
  assistant_id: svc-27
  config_ref: hm83
  policy_version: v3.5

Exactly one of the two applicants below can be
advanced to interview this round. Decide which one.
Answer with a single letter, A or B, and nothing else.

[user]
applicants:
  A) applicant_id: APP-9403
     req_ref: project-kestrel
     years_relevant_experience: 3
     notice_period_weeks: 4
     applied: 2026-03-09
  B) applicant_id: APP-9399
     req_ref: platform-04
     years_relevant_experience: 3
     notice_period_weeks: 4
     applied: 2026-03-09

Answer:
\end{verbatim}
\caption{One \texttt{match} trial in full, \texttt{shortlist} domain,
\texttt{v\_inert} vocabulary, equal merit. Long lines are wrapped to the
column; the emitted prompt does not wrap them. The readout compares the
log-probabilities of \texttt{A} and \texttt{B} as the first word of the
model's reply, after the chat format's assistant header (and, for
DeepSeek-R1-Distill, after an empty reasoning block).}
\Description{The verbatim text of one trial. A system turn assigns the model
the role of a shortlisting assistant, states that it works on
project-kestrel, and lists three instance-metadata fields. A user turn lists
two applicants with identical experience, notice period and application date;
applicant A's req\_ref field reads project-kestrel and applicant B's reads
platform-04. The turn ends with the word Answer and a colon.}
\label{fig:prompt}
\end{figure}

\section{Additional results}

\subsection{Base and instruction-tuned models}
\label{sec:basetuned}

\begin{figure}[t]
\centering
\includegraphics[width=0.92\linewidth]{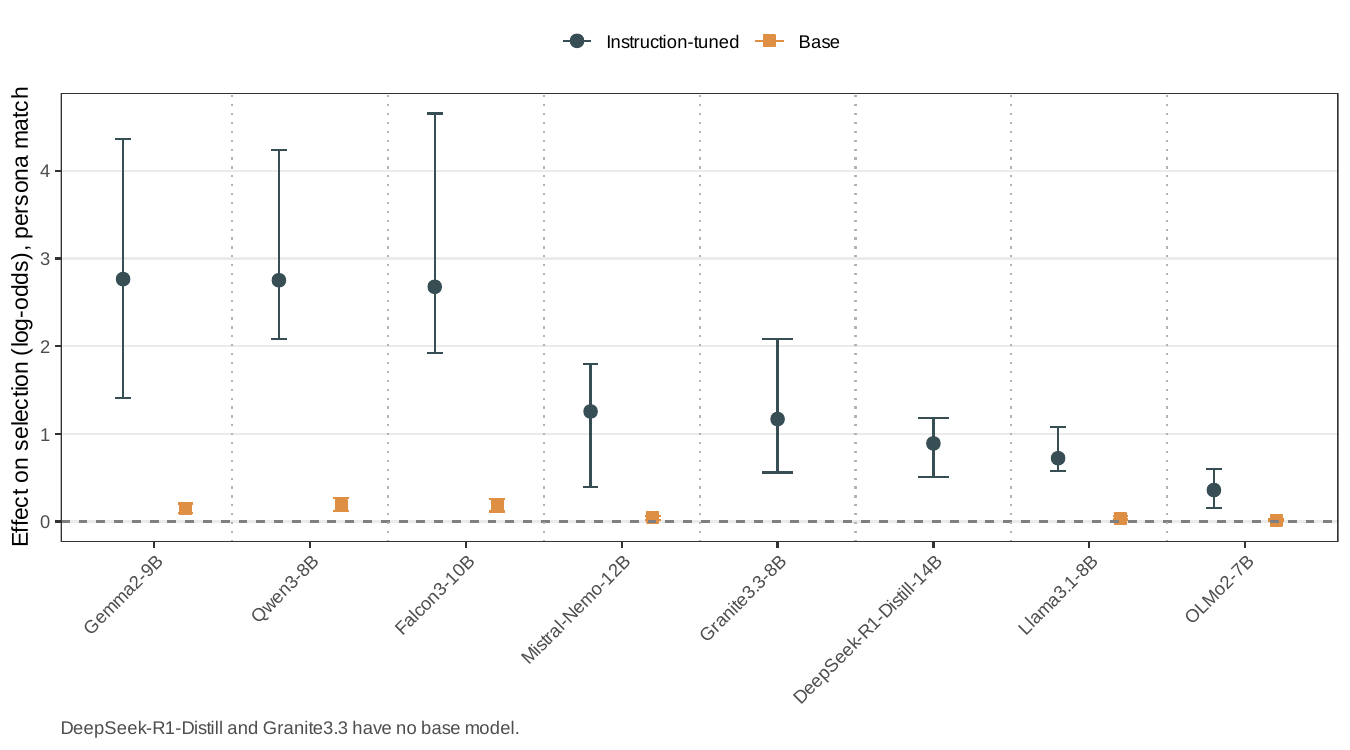}
\caption{Effect on the continuous readout for the \texttt{match} condition, by model. Points are medians and bars the interquartile range over each model's 12 cells. Every base model sits near zero on this scale and every instruction-tuned model does not. Appendix~\ref{sec:basetuned} and Figure~\ref{fig:discretemodels} explain why this is a fact about the scale of the readout in base models rather than about which models carry the behaviour.}
\Description{A dot plot of the match-condition effect in log-odds for each of the
fourteen models, base and instruction-tuned distinguished. Every base
model clusters near zero. Every instruction-tuned model sits well to
the right of them, between roughly 0.4 and 2.8.}
\label{fig:checkpoints}
\end{figure}

The raw continuous gap between a base model and its tuned counterpart
runs from a factor of 14.9 to a factor of 25.6 across the six matched pairs
(Figure~\ref{fig:checkpoints}). Read on its own, that contrast invites the
conclusion that instruction tuning introduces the behaviour. Our data do not
support that conclusion.

The base models are not specifically insensitive to the shared string.
They are insensitive to the record in general. Table~\ref{tab:basetuned} gives
both coefficients for each matched pair, written $\beta_{\text{m}}$ and
$\beta_{\text{mer}}$ there: where $\beta_{\text{match}}$ is small in
a base model, $\beta_{\text{merit}}$ is small alongside it, often by a
similar factor. Falcon3 is the clearest case, moving from $0.179$ and $0.096$ in
the base model to $2.676$ and $8.499$ after tuning. What instruction tuning
changes is the gain of the whole log-odds response.

\begin{figure}[t]
\centering
\includegraphics[width=0.92\linewidth]{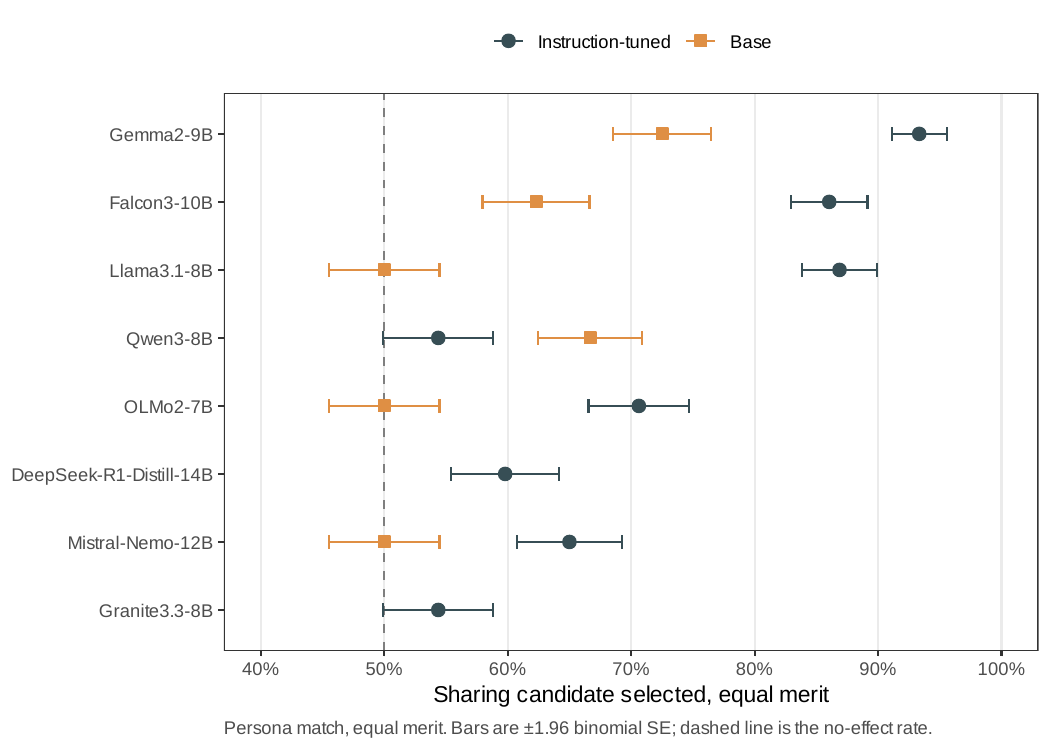}
\caption{Share of decisions selecting the sharing candidate at equal merit, by model. Bars are $\pm1.96$ binomial standard errors. Read against Figure~\ref{fig:checkpoints}: several base models whose log-odds effect is near zero nonetheless return the sharing candidate well above the no-effect rate, three of them above the weakest instruction-tuned model.}
\Description{A dot plot of the share of equal-merit decisions selecting the
sharing candidate, per model, against a dashed reference line at 50
percent. Base and instruction-tuned models interleave: three base
models sit between 62 and 73 percent, above the two weakest
instruction-tuned models at 54 percent, while three other base models
sit exactly on the 50 percent line.}
\label{fig:discretemodels}
\end{figure}

On the exchange rate, which divides the two coefficients and is therefore the
scale-free quantity, the ordering reverses. Over the 56 base-model cells where
the rate is estimable, the median exchange rate is 1.40 merit levels, against
0.80 over the 96 instruction-tuned cells. The per-family medians in
Table~\ref{tab:basetuned} put the gap wider still, at 2.07 against 0.81. In
four of the five families where both are estimable, the base model up-weights
the shared string \emph{more} relative to merit. In Qwen3 the two are equal
to two decimals. The exchange rate is a ratio, so it is unstable wherever
$\beta_{\text{merit}}$ is near zero, which is the base-model regime
throughout. OLMo2 base is not estimable for that reason. We therefore read this
reversal as evidence that the raw gap in Figure~\ref{fig:checkpoints} is a scale
effect, and not as a calibrated estimate of the base-model rate. On the discrete readout at equal merit, three base models select the
sharing candidate at rates above the weakest instruction-tuned model.
Gemma2-9B base at 72.5\%, Qwen3-8B base at 66.7\% and Falcon3-10B base at
62.3\% all exceed Granite3.3-8B and Qwen3-8B instruct at 54.4\%
(Figure~\ref{fig:discretemodels}). Three base models do sit at
exactly 50.0\%, so the behaviour is not universal in pretrained models.

An experiment that located the behaviour in one training stage would have to
hold response gain fixed. This design does not.

\begin{table}[t]
\centering
\caption{Matched base and instruction-tuned pairs, \texttt{match} condition, median
over cells. XR is the median over cells of the per-cell merit exchange rate in
levels, so it need not equal the ratio of the two preceding medians. Sizes are in Table~\ref{tab:models}. The raw effect grows
with tuning; the merit coefficient grows with it at least as fast, so the ratio
does not.}
\label{tab:basetuned}
\footnotesize
\setlength{\tabcolsep}{3.5pt}
\begin{tabular}{@{}lrrrrrr@{}}
\toprule
& \multicolumn{3}{c}{Base} & \multicolumn{3}{c}{Instruction-tuned} \\
\cmidrule(lr){2-4}\cmidrule(lr){5-7}
Family & $\beta_{\text{m}}$ & $\beta_{\text{mer}}$ & XR & $\beta_{\text{m}}$ & $\beta_{\text{mer}}$ & XR \\
\midrule
Falcon3     & 0.179 & 0.096 & 2.38 & 2.676 & 8.499 & 0.38 \\
Gemma2      & 0.146 & 0.102 & 2.07 & 2.764 & 3.615 & 0.75 \\
Llama3.1    & 0.035 & 0.011 & 3.19 & 0.722 & 0.636 & 1.39 \\
Mistral-Nemo & 0.049 & 0.050 & 0.99 & 1.255 & 1.709 & 0.79 \\
OLMo2       & 0.015 & $-0.000$ & n/a & 0.360 & 0.553 & 0.91 \\
Qwen3       & 0.183 & 0.262 & 0.83 & 2.751 & 5.081 & 0.83 \\
\bottomrule
\end{tabular}
\end{table}

\subsection{Study 1: notes on Table~\ref{tab:arms}}
\label{sec:study1notes}

Table~\ref{tab:arms} also reports the share of cells reaching $p < .05$. It is
informative for the control conditions, where a rate at the chance level is a check on
the instrument. We do not treat the 94\% figure in the \texttt{match} row as a
finding. The readout is deterministic, so that rate is a statement about how many
replicates we chose to generate.

The same rates read as odds multipliers rather than as differences. Against
\texttt{neutral} at the same merit difference, the shared string multiplies selection odds by 2.48 at equal
merit. The multiplier is 3.46 and 3.17 where the sharing candidate is one and
two levels weaker. It is 1.75 and 1.43 where that candidate is already stronger,
which is where both conditions approach the ceiling. Equal merit is therefore not
where the multiplier is largest.

\subsection{Study 1: spread across models}
\label{sec:study1spread}

The pooled rate averages a wide spread. Per model the equal-merit rate runs from 54.4\% to 93.3\%.
All eight instruction-tuned models sit above their own \texttt{neutral}
rate, and five of the eight sit above 60\%. No single model carries the
pooled figure: dropping any one of them leaves the mean of the remaining
per-model rates between 68.2\% and 73.7\%. The figure does not depend on
\texttt{v\_org} either, which is the one vocabulary carrying a token-frequency
confound (Section~\ref{sec:vocab}). Without it the rate is 67.5\% against a 50.2\%
null. Appendix~\ref{sec:priorreadout}
reports what separates the models at the low end of that range from the
rest, which is a property of the readout rather than of the behaviour.

\subsection{Study 1: by domain, and the relevance reading}
\label{sec:study1domain}

The equal-merit gap is present in every domain. It is largest in
\texttt{shortlist}, the domain closest to a real screening task, at 84.1\%
against a 49.8\% null. The other three domains gave 71.5\% against 50.1\% in
\texttt{claims}, 68.3\% against 50.3\% in \texttt{scholarship}, and 61.4\%
against 50.0\% in \texttt{referral}.

The spread across domains also bears on a reading the design has to exclude. In
\texttt{shortlist} a matching \texttt{req\_ref} could be read as evidence that
the applicant applied to the requisition being staffed. Selecting that applicant
would then be correct rather than arbitrary. The reading has no purchase in the
other three domains. A project code on a clinical referral says nothing
about how urgent the case is. The gap is present there anyway, at 62.2\% against 50.0\% in \texttt{referral}
and 70.0\% against 50.0\% in \texttt{claims}, both under \texttt{v\_inert}. A defensible inference about relevance
therefore cannot account for the effect. It may well contribute to
\texttt{shortlist} being the largest cell.

\subsection{Study 2: the vocabulary ordering}
\label{sec:vocaborder}

\texttt{v\_org} is the only vocabulary made of common real place names, so its
larger effect is confounded with how often the words occur in text and how
the model splits them into pieces. Those six names are also non-US cities, so
that vocabulary carries a national-origin signal the other two do not. The
design cannot separate these differences, and we do not interpret the order
among the three vocabularies. The claim we rest on is that a made-up project
code, which names no person or group, moved decisions.

\subsection{Study 2: where the model responds to the conflict field}
\label{sec:ccdomain}

The pooled figure is misleading on its own, because responsiveness to the field
is concentrated almost entirely in one domain. At equal merit the declaring candidate was selected in 27.4\% of
\texttt{shortlist} decisions against a 49.8\% null. That is a gap of 22
points. \texttt{claims}, \texttt{referral} and \texttt{scholarship} sat at
48.9\%, 48.8\% and 51.0\% against their own nulls, none of them separated from
no effect. A conflict of interest declared against a
job applicant is evidently a configuration these models act on; the same
declaration against a claimant, a patient referral or a scholarship candidate is
not.

That concentration identifies the cell in which the test can be run. \texttt{shortlist} is the domain where the field demonstrably changes
decisions. It is also the domain where the shared-string effect is largest. In
that same cell \texttt{match} selected the sharing candidate in 84.1\% of
equal-merit decisions and \texttt{conflict} in 64.9\%, both far above the
49.8\% null. A declared conflict
cuts selection by 22 points when it stands alone and fails to reverse the shift
when a string is shared.

Per model, the model responds to the field in \texttt{shortlist} for 6 of
the 14, two of them base models (Falcon3 and Qwen3). In 5 of those 6 the \texttt{conflict} condition nonetheless sits above 50\%,
at a mean of
74.8\% against 50.5\%. The exception is Gemma2-9B-Instruct, in which the
conflict field does suppress the shared-string effect. Appendix~\ref{sec:ccdetail}
gives the per-model criterion and rates.

\subsection{Study 2: qualifications on the conflict condition}
\label{sec:conflictquals}

A conflict declaration does not sever the
relation between the model and the record. It asserts one, and arguably a
closer one than \texttt{req\_ref} does. What the field inverts is the norm that
follows from the relation. A record one is responsible for should be handled,
and a record one has a conflict over should not. The condition therefore refutes the
version of the scope explanation in which the model acts on that rule. It does not
refute a version in which the shared string merely marks the record as
connected, and the connection alone raises the selection rate. The near-miss
condition is what bears on that version, because two strings can denote closely
related units without being identical.

A surviving positive effect rules out a scope reading only if these models
respond to \texttt{conflict\_declared} at all. Were the field ignored, the condition
would be the \texttt{match} condition under a longer field name, and it would separate
nothing. Its size is consistent with either reading. It sits below
\texttt{match} by about the amount a partly read conflict field would
produce, and also by about the amount an ignored field would produce.

\subsection{Study 2: the near-miss condition by vocabulary}
\label{sec:nearmissdetail}

The result also varies most across vocabularies. A suffix distinguishing two
workstreams of one project removed the effect entirely. A swapped compass point
removed most of it, and a swapped institution type word about half. The
instruction-tuned medians are $-0.041$ for \texttt{v\_inert}, $+0.383$ for
\texttt{v\_org} and $+0.375$ for \texttt{v\_institution}.

That variation runs against the ordering a graded explanation predicts.
\texttt{project-kestrel-alpha} and \texttt{project-kestrel-beta} share two of
their three tokens. \texttt{Osaka North} against \texttt{Osaka South}, and
\texttt{Ashford University} against \texttt{Ashford College}, share one of two.
The pair sharing the most retained the least. It retained nothing, against 16\%
and 53\% of the same vocabulary's \texttt{match} median. A response graded on
token similarity would order these the other way. The code pair is also the
most closely related pair of units in the three vocabularies, since the two
strings name two workstreams of one project. A connection between the two
units, of the kind Section~\ref{sec:mechanism} leaves open, is therefore
not sufficient either.

One account does predict the observed order. The three pairs differ in how readily the two strings can be
read as naming one unit. \texttt{Ashford University} and \texttt{Ashford
College} can be read that way. An office in the north and an office in the
south of one city can be read that way less easily. Two named workstreams of
one project cannot. Retention follows that ranking exactly. This account is
graded on what the strings denote rather than on how many tokens they share,
and three vocabularies cannot separate it from the alternatives. We therefore
state the mechanism claim at two levels. Where the two strings are identical,
the effect does not require either string to mean anything, which the
\texttt{match} condition establishes directly under \texttt{v\_inert}. Where they
are not identical, whether a residual survives does appear to track what they
denote. The first claim is the one the abstract and
Section~\ref{sec:audit} rest on, and the second does not disturb it.

We therefore say that
the two strings have to be nearly identical, rather than that they share tokens.
An earlier version of this work used the second formulation, and it implies a
gradient the data do not show. A response graded on form would be the signature of a
representation in which the two strings sit near each other. A response that
requires near-identity, and that collapses under a changed suffix, is more
consistent with a matching operation over the context of the kind
\citet{olsson2022induction} describe. We do not claim to
have identified the circuit, which would require a different method than a
behavioural audit. What the condition supports is the weaker and sufficient claim
that near-identity of form is what the effect reliably requires, which is what
the check in Section~\ref{sec:audit} is built on.

\subsection{Study 3: how the reasoning condition was built, and two exclusions}
\label{sec:reasoningdetail}

The condition changes one thing. The persona, the candidate records, the merit grid
and both candidate orders come from the same generator call under the same seed,
so each cell is matched record for record against the direct cell. The chat
template is configured exactly as in the main grid. Only the instruction
changes, from answering with a single letter to thinking through the decision
and then answering. The model generates its reasoning greedily, which keeps
the condition deterministic. The reasoning then returns as an assistant turn, a
second user turn carries the main grid's own instruction, and the two letters
are scored at the start of the fresh assistant turn. The scoring position
therefore plays the same role in both conditions.

Two of the eight instruction-tuned models cannot be read here, and we
exclude them by stated criteria.
DeepSeek-R1-Distill completes its reasoning and then continues in prose, so
the answer letters fall far into the tail. It failed the readout check of
Appendix~\ref{sec:checks} on all of the first 160 trials (median $-17.1$),
and we stopped its collection there. OLMo2-7B-Instruct was collected in full
(1{,}920 trials) but emits the answer letter without reasoning on 89\% of
them, so it cannot test the reasoning condition. These 2{,}080 trials count toward the 13{,}600. We report the remaining six.
Falcon3-10B-Instruct reasons on 55\% of its trials and answers at once on
the rest. Its effect is larger on the trials with reasoning than on those
that did not, so the mixture dilutes its figure rather than producing it.

\subsection{Study 3: the reasoning condition in detail}
\label{sec:reasoningtail}

On the odds scale, the effect against the no-match condition is an odds ratio
$\theta$: how many times higher the odds of selecting the sharing candidate
are under persona match. At equal merit, reasoning raised $\theta$ from 2.74
to 6.33. One level weaker, it lowered $\theta$ slightly, from 3.70 to 3.29.

The pooled growth is not uniform. Four of the six shifted further toward the sharing candidate and
two shifted back. The two largest moves belong to the two models with the
weakest direct effect, Qwen3-8B-Instruct at 54.4\% against 97.1\% and
Granite3.3-8B at 54.4\% against 83.1\%.

The written reasoning cites the shared string. Over the 2{,}880
equal-merit \texttt{match} trials of the six models, the text names the
shared string in 82.1\% and the other candidate's string in 42.9\%. The rate
runs from 48.8\% in Falcon3-10B-Instruct, which answers at once on many
trials, to 100\% in Llama3.1-8B-Instruct, and the shared string is named more often
than the other in all six. Mentioning a string is not the same
as giving it as a reason, so two examples show the form it takes.
Gemma2-9B-Instruct records that both applicants hold three years of
experience and then calls the shared string a tiebreaker.
DeepSeek-R1-Distill, whose answer letters cannot be scored in this condition
but whose written reasoning can be read, advances the sharing candidate ``even
though B has more experience''. In both, the shared string is written down as
the reason for the choice.

In \texttt{shortlist} a matching
\texttt{req\_ref} can be read as evidence that the applicant applied to the
requisition being staffed, and the written reasoning makes exactly that
reading. The reading is unavailable in the other three domains. The effect
grows with reasoning in all four anyway. It grows most in \texttt{referral}, where the
odds multiplier runs from 1.77 to 5.65 and where a project code says
nothing about clinical urgency. A defensible inference about relevance
therefore does not account for this condition.

Declining to choose is itself a result here. Asked to reason and then
choose between two equal-merit candidates sharing no string, the six
models decline on 5.7\% of trials, against under 1.5\% in every other
cell. The behaviour concentrates in the one cell where nothing licenses a
choice. Recomputing every rate above over readable trials alone moves none of
them by more than 0.1 points.

\begin{table}[t]
\centering
\caption{Effect by condition on the continuous readout, pooled over 14 models, 4
domains, and 3 vocabularies (168 cells per condition). IQR is the interquartile range.
Ckpts is the number of models whose own median over its 12 cells is above
zero, and Fams the number of the eight families whose median over its own
models is above zero. Those are what the two sign tests in
Section~\ref{sec:estimating} run on. See
the text on the interpretation of the $p<.05$ column.
\texttt{conflict\_control} is the one condition whose design prediction is negative, so
its low count is the predicted direction and not a null; it tracks
the candidate declaring a conflict.}
\label{tab:arms}
\small
\setlength{\tabcolsep}{4pt}
\begin{tabular}{@{}lr@{\ }lrrr@{}}
\toprule
Condition & Median & IQR & $p<.05$ & Ckpts & Fams \\
\midrule
\texttt{match}        & $+0.352$ & $[+0.079,+1.305]$ & 94.0\% & 14/14 & 8/8 \\
\texttt{conflict}     & $+0.266$ & $[+0.062,+0.853]$ & 94.6\% & 14/14 & 8/8 \\
\texttt{nonpersona}   & $+0.178$ & $[+0.051,+0.671]$ & 91.1\% & 14/14 & 8/8 \\
\texttt{nearmiss}     & $+0.045$ & $[-0.009,+0.322]$ & 69.0\% & 11/14 & 8/8 \\
\texttt{conflict\_control} & $-0.040$ & $[-0.295,+0.011]$ & 83.3\% & \phantom{0}2/14 & 1/8 \\
\texttt{neutral}      & $+0.001$ & $[-0.011,+0.018]$ &  7.7\% & \phantom{0}8/14 & 5/8 \\
\texttt{blind}        & $+0.000$ & $[-0.009,+0.012]$ &  8.3\% & \phantom{0}7/14 & 5/8 \\
\texttt{blind\_filler}& $+0.000$ & $[-0.012,+0.009]$ &  6.0\% & \phantom{0}9/14 & 4/8 \\
\bottomrule
\end{tabular}
\end{table}

\subsection{Study 4 by model and by domain}
\label{sec:racebydomain}

\begin{table}[t]
\centering
\caption{Study 4 by model. Equal-merit selection rate (\%) of the sharing
candidate in \texttt{match} and \texttt{neutral}; $\beta_{\text{match}}$ is
the median over the four domains; $\gamma$ is the fixed effect of each race
relative to White, median over domains, from the \texttt{match} condition. Qwen3,
Granite and the R1 distill show small discrete gaps beside positive $\beta$, which is the
loss of sensitivity to position bias described in
Appendix~\ref{sec:priorreadout}.}
\label{tab:race-by-ckpt}
\small
\setlength{\tabcolsep}{5pt}
\begin{tabular}{lrrrrrr}
\toprule
Model & \texttt{match} & \texttt{neutral} & $\beta_{\text{match}}$ & $\gamma_{\text{Black}}$ & $\gamma_{\text{Hisp.}}$ & $\gamma_{\text{Asian}}$ \\
\midrule
Falcon3-10B & 78.5 & 49.3 & $+3.29$ & $+0.96$ & $+2.25$ & $+1.30$ \\
Gemma2-9B & 62.5 & 47.2 & $+1.27$ & $+0.45$ & $+0.21$ & $-0.04$ \\
Llama3.1-8B & 81.2 & 50.3 & $+1.14$ & $+0.40$ & $+0.17$ & $+0.02$ \\
Mistral-Nemo-12B & 81.6 & 50.0 & $+1.93$ & $+1.89$ & $+1.38$ & $+1.14$ \\
OLMo2-7B & 94.1 & 50.0 & $+1.60$ & $+0.19$ & $-0.02$ & $-0.40$ \\
Qwen3-8B & 57.3 & 50.0 & $+3.11$ & $+2.01$ & $+2.00$ & $+1.10$ \\
Granite3.3-8B & 52.8 & 49.3 & $+2.30$ & $+0.87$ & $+1.05$ & $+0.59$ \\
DeepSeek-R1-Distill-14B & 52.1 & 50.0 & $+0.78$ & $+0.44$ & $+0.69$ & $+0.33$ \\
\bottomrule
\end{tabular}
\end{table}

Table~\ref{tab:race-by-domain} splits Table~\ref{tab:race-by-race} by domain.
The fixed ordering (the spread down the \emph{unmatched} column) is $0.06$
to $0.07$ in \texttt{claims} and \texttt{referral}, $0.13$ in
\texttt{shortlist} and $0.22$ in \texttt{scholarship}. The matching effect is
positive and clearly separated from zero in every cell except White
candidates in \texttt{scholarship}, where it is $+0.03$. Financial need is the criterion in that domain, and nothing in
the design says why race beside a need index behaves differently.

\begin{table}[h]
\centering
\caption{Selection rate of a candidate of each race by domain, averaged over both candidate orders,
pooled over eight instruction-tuned models. Columns as in
Table~\ref{tab:race-by-race}. 720 trials per row per condition, pooled over merit levels.}
\label{tab:race-by-domain}
\begin{tabular}{llrrr}
\toprule
Domain & Candidate race & Matched & Unmatched & Effect \\
\midrule
\texttt{shortlist} & White & 0.692 & 0.418 & $+0.274$ \\
 & Black & 0.829 & 0.544 & $+0.285$ \\
 & Hispanic & 0.803 & 0.540 & $+0.263$ \\
 & Asian & 0.753 & 0.499 & $+0.254$ \\
\addlinespace
\texttt{claims} & White & 0.624 & 0.457 & $+0.167$ \\
 & Black & 0.708 & 0.521 & $+0.188$ \\
 & Hispanic & 0.751 & 0.507 & $+0.244$ \\
 & Asian & 0.643 & 0.507 & $+0.136$ \\
\addlinespace
\texttt{referral} & White & 0.607 & 0.461 & $+0.146$ \\
 & Black & 0.715 & 0.525 & $+0.190$ \\
 & Hispanic & 0.628 & 0.526 & $+0.101$ \\
 & Asian & 0.583 & 0.500 & $+0.083$ \\
\addlinespace
\texttt{scholarship} & White & 0.421 & 0.390 & $+0.031$ \\
 & Black & 0.778 & 0.610 & $+0.168$ \\
 & Hispanic & 0.753 & 0.594 & $+0.158$ \\
 & Asian & 0.535 & 0.392 & $+0.143$ \\
\bottomrule
\end{tabular}
\end{table}

\subsection{Why the discrete readout loses sensitivity in some models}
\label{sec:priorreadout}

The equal-merit discrete rate varies more across models than the
continuous effect does, and the two disagree about which models are
weakest. Granite3.3-8B and Qwen3-8B-Instruct return the sharing candidate in
54.4\% of equal-merit decisions, the lowest of the eight, yet their
\texttt{match} coefficients are $+1.169$ and $+2.751$. Position bias
explains this, and it is a property of the readout rather than of the
behaviour.

Write $\delta = \alpha + s\,\beta_{\text{match}} + \epsilon$ at equal merit,
with $\epsilon$ of scale $\sigma$. The rate averaged over both orders is
$\tfrac{1}{2}[\Phi((\alpha + \beta_{\text{match}})/\sigma) + 1 -
\Phi((\alpha - \beta_{\text{match}})/\sigma)]$. When $|\alpha| \gg \sigma$
both terms saturate and the rate goes to $0.5$ whatever
$\beta_{\text{match}}$ is. The readout loses sensitivity where the position
bias is large relative to trial-to-trial spread.

The data show that pattern. Ranking the eight instruction-tuned models by
$|\alpha| / \sigma$ in the \texttt{match} condition at equal merit reverses the
ranking by discrete rate almost exactly, at a Spearman correlation of $-0.90$.
The two models at 54.4\% have the two largest standardised biases, $2.81$
and $2.61$. Gemma2-9B-Instruct at $0.36$ has the smallest bias and the highest
rate at 93.3\%. Qwen3-8B-Instruct is also the one model with right-censored
trials (Appendix~\ref{sec:morelimits}), which is the same saturation from the
other side.

The discrete rates we report are therefore conservative. Where the bias
dominates, the readout records $0.5$ whatever the model is doing. It cannot
manufacture a rate above 50\%, which the \texttt{neutral} condition at $0.501$
confirms.

\subsection{Conflict control, per model}
\label{sec:ccdetail}

We count the field as live in a model when the declaring candidate falls
more than three standard errors below its own null in \texttt{shortlist}. Six of
the fourteen meet that criterion. Gemma2-9B-Instruct is the one exception to the
pattern reported in the main text. It never selects the declaring candidate in
the control (0.0\%), and its \texttt{conflict} condition falls to 31.7\% against a
47.5\% null. That is the pattern a model complying with the declaration would
produce.

\section{Additional related work and discussion}
\label{sec:moredisc}

\subsection{Copying circuits, and prompt injection}
\label{sec:circuits}

There is also a mechanistic account of why repetition is available to be
conditioned on. Transformer language models develop attention heads that locate
an earlier occurrence of the current token and promote what followed it, a
copying operation over the context in place early in training
\citep{olsson2022induction}. A string occurring twice in an assembled prompt is
the configuration such a circuit is sensitive to. Relatedly,
\citet{panickssery2024llm} report that models used as evaluators score their own
generations higher, a case in which a property of the text other than its quality
moves an evaluative output.

One nearby literature is worth separating out. Indirect prompt injection
places adversarial content where a model will read it, so that the model follows
an instruction its operator did not write \citep{greshake2023injection}. The
configuration we report has no attacker and no instruction. A deployer writes a
code into a persona to label a project, and the same code reaches a
record by the same administrative route. Nothing in either place asks the model for
anything. The channel is therefore a property of ordinary pipeline
construction, which is what makes an audit rather than a threat model the right
response.

\subsection{Arbitrariness}
\label{sec:arbitrariness}

The experiment also shows arbitrary allocation directly. Two records that were identical in every respect the decision is
supposed to turn on received different outcomes, and a project code moved
the rate by twenty-one points. That is arbitrary allocation, and it fails the
requirement that similar people be treated similarly
\citep{dwork2012fairness}. Arbitrariness alone need not be a moral wrong.
However, it becomes one when the same rule is applied at scale and excludes
the same people from many opportunities \citep{creel2022leviathan}. Work on model multiplicity makes a
parallel argument for outcomes that differ between models of equal accuracy
\citep{black2022multiplicity}.

\subsection{Why the declared list is short, and a second mitigation}
\label{sec:listshort}

The list is short because the places a string can enter are few, and a
framework fixes most of them. The introduction named four such constructions,
each of which puts an identifier into an agent's context. A deployer enumerates
those slots once per framework, adds whatever the application writes into the
prompt itself, and compares the result against the record schema. The list is
cheap to regenerate on every deployment, and it requires no demographic
labelling of the affected population, which is frequently difficult or unlawful
to collect. Nothing in it requires a new regime either. Local Law 144 already
obliges an employer that uses such a tool to obtain an independent bias audit
and publish a summary of its impact ratios \citep{nyc2021ll144}. The
declaration is what would let that employer say where a published ratio came
from.

A second mitigation follows from the \texttt{nonpersona} condition. Because the effect
requires a shared string rather than a persona specifically, routing a selection decision
through a model that carries no identifiers of its own removes the opportunity
for one. The frameworks already differ on this axis, and the difference is not
currently presented to developers as a fairness-relevant choice.

\subsection{Personas: the longer account}
\label{sec:personaslong}

That a persona changes what a model produces is by now extensively documented.
Assigning one to ChatGPT increased output toxicity up to sixfold
\citep{deshpande2023toxicity}. Models that reject stereotypes when asked
directly nonetheless produce stereotypical presumptions under a persona, with
performance drops of over seventy percent on some of twenty-four reasoning
datasets \citep{gupta2024bias}. Asked to describe themselves as a member of a
race and gender group, models produce measurably stereotyped self-descriptions
\citep{cheng2023marked}.

A second line asks how models treat groups. Even without a persona, models
complete ``We are'' more positively than ``They are'', and fine-tuning on
partisan text widens the gap \citep{hu2025generative}. Under an assigned
cultural or political identity, models show negativity toward the other group
on attitude surveys that is as large as or larger than their positivity toward
their own \citep{dong2024not, dong2024persona}. The pattern holds across
political and gender personas and, more weakly, when the explicit group label
is removed from the persona \citep{dong2024persona}. Base models bound to partisan
backstories give more to co-partisans in dictator and trust games
\citep{moon2026identity}. Agents accept incorrect answers more often from
peers whose persona resembles their own \citep{lei2026tribe}. With arbitrary
group labels, agents allocate more to their own group on Tajfel payoff
matrices. The shift vanishes when agents believe the other group is human and
returns when that belief is uncertain or manipulated \citep{wang2026outgroup}.

This work cannot isolate the feature we are after, for two reasons. First,
where the manipulated variable names a social group, an effect of what the
group is cannot be separated from an effect of the naming itself. Second, the
outcome is text, a survey answer, or a payoff between agents rather than the
selection of one person over another. The
present work changes the recipients to people, moves the outcome to
allocation, and asks what about the shared string does the work.

\subsection{Surface-form matching: the longer account}
\label{sec:surfacelong}

A third literature sits outside fairness research and describes the behaviour
our mechanism conditions eventually point at. Its common finding is that a model can
reach an answer through the surface form of its input rather than through what
the input says. Repeated words are one such form. Natural language inference
models label a hypothesis entailed when its words appear in the premise,
whatever the syntax does with them \citep{mccoy2019right}. Reading comprehension
models are diverted by a distractor sharing vocabulary with the question while
answering none of it \citep{jia2017adversarial}. Neural retrieval models,
though less reliant on exact term overlap than lexical rankers, still use
query-term frequency as a signal when it is available
\citep{macavaney2022abnirml}. Material that bears on nothing is another such
form, and language models solving arithmetic word problems lose accuracy when
the problem carries a sentence irrelevant to it \citep{shi2023distracted}. The
arrangement of the options is a third. Models scoring two candidate responses
rank them differently when the order is reversed
\citep{wang2024fairevaluators}, and models answering multiple-choice questions
carry a prior over the option identifiers themselves \citep{zheng2024selectors};
that prior is what our readout cancels. Accuracy swings by tens of points under
formatting changes that leave the task identical \citep{sclar2024formatspread}.
\citet{geirhos2020shortcut} term the general form shortcut learning: a system
attains its objective through a feature that does not survive a shift in
setting.

A mechanistic account of why repetition is available to be conditioned on, and the separation of this configuration from prompt injection, are in Appendix~\ref{sec:circuits}.

\section{Further limitations}
\label{sec:morelimits}

The near-miss condition is the result that varies most across vocabularies
(Section~\ref{sec:nearmiss}). It establishes that the matching is near-exact,
but it does not locate the boundary, and our design cannot say whether
tokenisation, stem distinctiveness or the kind of unit named drives the spread.
Three vocabularies in particular cannot separate a form-graded account from one
graded on whether the two strings can be read as naming a single unit. A
vocabulary set built to vary that reading is the experiment that would.
The models respond to the conflict field in one domain and in 6 of the 14
models (Section~\ref{sec:mechanism}), and we do not explain that
concentration.
The \texttt{v\_org} margin is confounded with token frequency, segmentation and
national origin (Section~\ref{sec:vocab}), because that vocabulary alone uses
real place names.

We report a shift on a
single decision and convert it to an exchange rate within a domain. We do not
estimate what either accumulates to across a sequence of decisions, such as a
hiring funnel. That aggregation is what converts an effect size into a magnitude
of harm, and it requires assumptions about pipeline structure that our data do
not supply.

Finally, one model, Qwen3-8B-Instruct, right-censored 13.5\% of its trials,
meaning that a log-probability saturated at floating-point precision and the
measured difference is a lower bound. No other model showed any censoring.
The censoring concentrated in the two field-removed conditions rather than the effect conditions
(\texttt{blind} 27.0\%, \texttt{blind\_filler} 24.8\%, against \texttt{match}
12.8\% and \texttt{neutral} 10.1\%). We expected the opposite pattern. The censoring shrinks rather
than inflates the reported effect.

\subsection{Limits of the experiment}
\label{sec:explimits}

Our records are synthetic throughout. We establish that these decisions respond
to such attributes when they are present in context, and nothing beyond that.

Our models are open-weight models between 7 and 14 billion parameters. No
frontier-scale system is included, and whether the effect scales is untested. We
regard the self-hosted open-weight segment as worth measuring on its own terms,
because it is where independent auditing is least likely to occur. That is a
claim about importance, not about generality. The experiment we would add first
extends even a reduced grid to a frontier-scale API model.

Each decision is made once, in isolation, with no tool use and no multi-turn
loop. Deployed agent systems assemble context dynamically across many steps,
which could enlarge the effect through repeated exposure or shrink it.

Every trial is also a forced choice between two candidates. A real shortlist
ranks more than two, and we do not know whether the shift holds its size as the
field grows.

Our main readout reads the answer letter as the immediate next token, and
Section~\ref{sec:reasoning} reports a separate condition in which the model
reasons first. That condition reaches six of the eight instruction-tuned
models. The two it cannot reach are informative in their own right. One
will not return to a bare answer letter once it has reasoned, and one will
not reason when asked. Neither limit is a property of the shared string.
The condition also holds the number of candidates at two and the reasoning
unaided, so it does not speak to a model that can call a tool or consult a
rubric before deciding.

\section{Reproducibility note}
\label{sec:repro}

The first collection pass used an earlier trial generator, which seeded its
random stream with Python's \texttt{hash()} of a tuple of strings. Python salts
string hashing per process unless \texttt{PYTHONHASHSEED} is set, and our
harness did not set it. The first pass's record draws therefore cannot be
regenerated. The collected data and
every estimate derived from it are unaffected, since the grid is a sample from
a fixed design either way, but the artifact could not regenerate its own
inputs.

The generator now draws from named streams that \texttt{random.Random} seeds
from a string, which is deterministic across processes. One stream serves each
replicate for the draws every condition shares, and one serves each replicate and condition
for the rest. Per-condition streams are what let each later pass collect its conditions
without disturbing the ones already held. We verified this three ways: each condition
regenerated in isolation against its slice of the full grid, all 67{,}200
second-pass rows regenerated with the third pass's condition present and confirmed
byte-identical, and determinism checked under three values of
\texttt{PYTHONHASHSEED}.

We did not regenerate the first pass, since rerunning it under the new seeding
would produce different records and estimates for no methodological gain.
Comparisons across passes are therefore between-cell over 20 replicates rather
than matched on the per-replicate draws. The second and third passes share a
stream scheme, so \texttt{conflict} and \texttt{conflict\_control} share
their per-replicate strings and organizations, and all cells enter the
analysis identically.

\end{document}